\documentclass[letterpaper, 10 pt, conference]{ieeeconf}  

\IEEEoverridecommandlockouts                              

\usepackage{amsmath} 
\usepackage{amssymb}  
\usepackage{graphicx}
\usepackage{subfigure}
\usepackage{multirow}

\title{\LARGE \bf Scale jump-aware pose graph relaxation for\\monocular SLAM with re-initializations}

\author{Runze Yuan$^{1}$, Ran Cheng$^{2}$, Lige Liu$^{2}$, Tao Sun$^{2}$, and Laurent Kneip$^{1}$
\thanks{$^{1}$Mobile Perception Lab, ShanghaiTech University}%
\thanks{$^{2}$Midea Robozone}%
}

\begin{document}

\maketitle
\thispagestyle{empty}
\pagestyle{empty}


\begin{abstract}
Pose graph relaxation has become an indispensable addition to SLAM enabling efficient global registration of sensor reference frames under the objective of satisfying pair-wise relative transformation constraints. The latter may be given by incremental motion estimation or global place recognition. While the latter case enables loop closures and drift compensation, care has to be taken in the monocular case in which local estimates of structure and displacements can differ from reality not just in terms of noise, but also in terms of a scale factor. Owing to the accumulation of scale propagation errors, this scale factor is drifting over time, hence scale-drift aware pose graph relaxation has been introduced. We extend this idea to cases in which the relative scale between subsequent sensor frames is unknown, a situation that can easily occur if monocular SLAM enters re-initialization and no reliable overlap between successive local maps can be identified. The approach is realized by a hybrid pose graph formulation that combines the regular similarity consistency terms with novel, scale-blind constraints. We apply the technique to the practically relevant case of small indoor service robots capable of effectuating purely rotational displacements, a condition that can easily cause tracking failures. We demonstrate that globally consistent trajectories can be recovered even if multiple re-initializations occur along the loop, and present an in-depth study of success and failure cases.
\end{abstract}

\section{INTRODUCTION}

Exteroceptive sensor-based Simultaneous Localization And Mapping (SLAM) is a fundamental solution required in many robotics and intelligence augmentation scenarios. In comparison to its alternatives, the basic case of monocular visual SLAM is often considered a highly attractive solution owing to low cost, low weight, low energy demands, and low extrinsic calibration or sensor fusion requirements. However, using only a single camera often also poses an intricate challenge, which is the scale invariance of its geometry. The present paper focuses on the important SLAM back-end optimization problem of pose graph relaxation with a particular view onto its ability to handle the scale invariance occurring in the monocular setup.

\begin{figure}[t]
  \centering
  \includegraphics[width=\columnwidth]{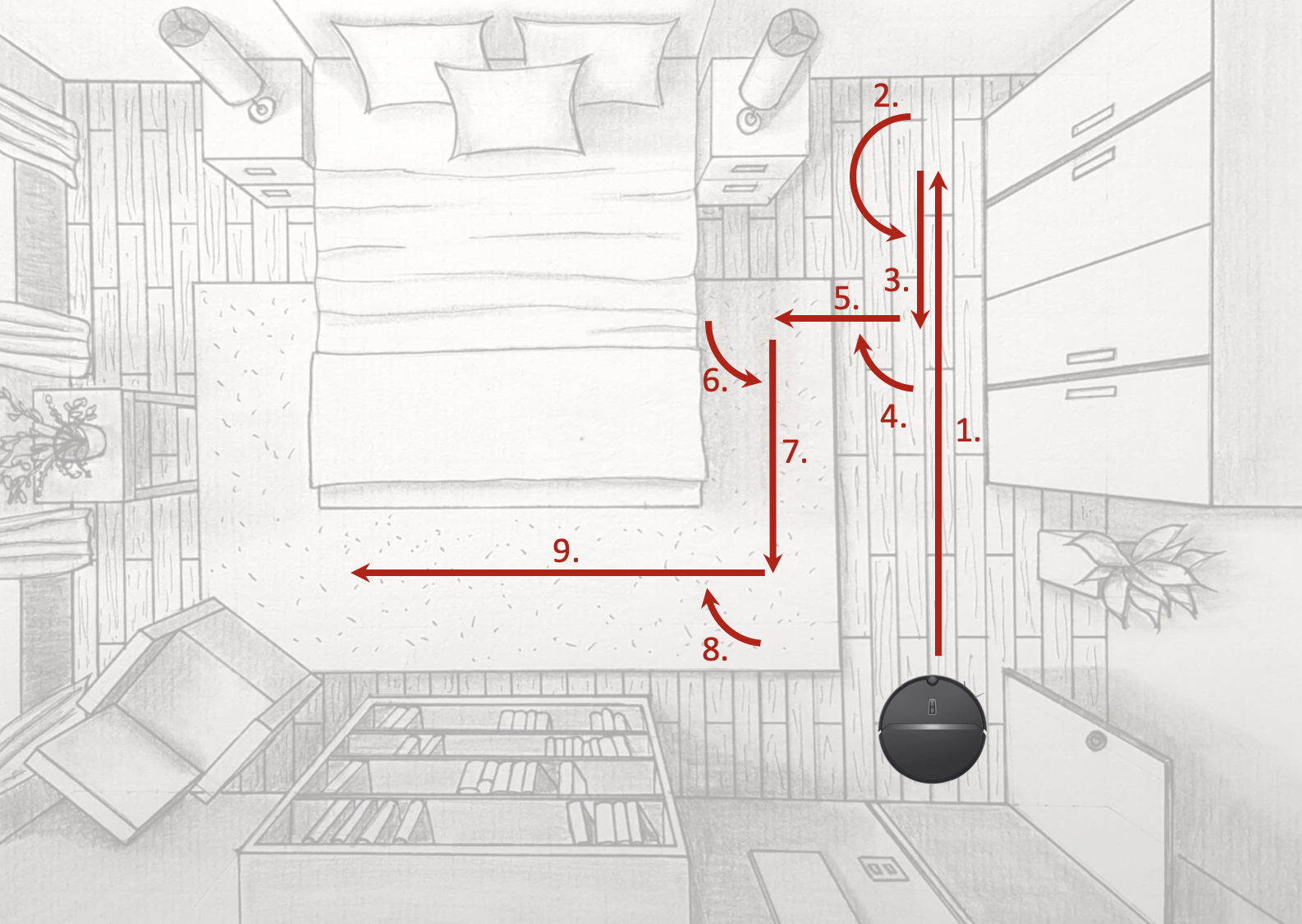}
  \caption{Illustration of the scenario addressed in this work. An agent equipped with a single monocular camera (e.g. a robotic vacuum cleaner) is traversing a room. Owing to the regular execution of purely rotational displacements, the monocular SLAM algorithm frequently loses tracking and needs to re-initialize. The result is a piece-wise scale-consistent trajectory. Our contribution is a hybrid pose graph optimization frame-work with the ability to reconcile a globally consistent trajectory.}
  \label{fig:frontFigure}
\end{figure}

Pose graph relaxation is a fundamental sub-problem of SLAM that aims at estimating a set of (often sequentially captured) sensor poses relative to a common, global coordinate system. The problem can be represented by a graph where the nodes correspond to the optimized sensor poses, and the edges represent measured relative positions and orientations between pairs of sensor frames. The goal of pose graph optimization is to find the optimal set of sensor poses that will minimize the overall discrepancy with the measured relative poses. Pose graph relaxation is often referred to as pose averaging as the measured relative poses may be inaccurate and lead to cycle inconsistencies. Hence, there typically are no absolute sensor poses that will simultaneously satisfy all relative pose constraints, and errors will have to be distributed over loops.

Applying pose graph optimization in the back-end of monocular visual SLAM requires special attention to the problem of scale invariance. The issue of a global scale invariance is easily addressed by simply estimating the euclidean sensor poses in the scaled domain, and---if the scale factor eventually becomes known---simply rescale the estimated sensor positions at the end of the optimization. However, the problem is that the scale factor is not globally consistent. Rather, it is a local variable and---owing to small error accumulations in each relative sensor pose estimation---subject to slow variation over time, a problem commonly referred to as \textit{scale drift}. Although the small scale change between successive frames is the result of estimation errors and thus cannot be determined (i.e. the relative scale can only be assumed to be one), scale change can be determined if a known place is revisited, an event known as a \textit{loop closure}. As introduced by Strasdat et al.~\cite{strasdat10}, expressing both absolute and relative sensor transformations as similarities therefore leads to scale drift-aware pose graph optimization.

While certainly an advancement, pose graph relaxation using similarity transformations still suffers from an important drawback, which is that the relative scale change between neighboring sensor frames either needs to be small to the point of being ignorable or otherwise measurable. However, in monocular SLAM there is an important situation in which neither of the two is true. If a tracking failure occurs, the local map may easily leave the field of view of the camera before a successful re-initialization occurred. A possible lack of overlap between the current and the previous local map then makes it impossible to reconcile the in principle arbitrary relative scale between the corresponding sensor frames. Rather than just being an occasional error, the discussed problem occurs systematically on dual-drive platforms that are equipped with a single perspective camera and occasionally exhibit purely rotational displacements (e.g. a robotic vacuum cleaner [cf. Figure~\ref{fig:frontFigure}]).

We make the following contributions:
\begin{itemize}
\item We demonstrate that the simple solution of setting the relative scale to one and applying scale drift-aware relaxation can lead to severe distortions in the case of scale jumps caused by re-initializations.
\item We introduce a novel hybrid pose graph optimization framework that admits scale-ignorant factors along re-initialization edges, and demonstrate how this seemingly simple change leads to scale reconciliation across loopy graphs and distortion-free estimates even if facing multiple tracking failures along a loop.
\item We present a complete discussion on the number and spatial configuration of failure edges that can be tolerated, and reveal critical configurations in which a globally consistent estimate (up to a single global scale factor) can no longer be obtained.
\end{itemize}

Our paper is organized as follows. Section~\ref{sec:relatedWork} reviews the related literature on pose graph optimization. Section~\ref{sec:theory} reviews both traditional loop closure as well as our modified, Hybrid Pose Graph Optimization (HPGO). Section~\ref{sec:experiments} finally presents the experimental validation of our claims using both simulated and real datasets.

\section{RELATED WORK}
\label{sec:relatedWork}

Pose graph optimization is a fundamental SLAM alternative to incremental filtering~\cite{davison07} and batch optimization over both poses and landmarks~\cite{triggs00}. Also known as graph SLAM, it limits the estimation to the sensor frame poses, thereby achieving efficient large-scale registration capabilities. Furthermore, pose graph optimization is more easily amenable to scenarios in which local constraints do not adhere to simple noise models (e.g. Gaussian distributions) and motion models may not be given.

The first approaches presented in the literature are restricted to the planar case and perform least-squares optimization using LU-decomposition~\cite{lu97}. Gutmann and Konolidge~\cite{gutmann99} later on apply the technique as a back-end module to reconcile front-end estimations coming from both incremental registration and loop closure detections. Later contributions mainly focus on the efficiency of the pose graph optimization by employing relaxations~\cite{howard01,duckett02,frese05} and local updating~\cite{howard01,duckett02,frese05,olson06}. The seminal contribution of Olson et al.~\cite{olson06} leads to a significant improvement of optimization robustness and efficiency using tree-based updates~\cite{frese06} and stochastic gradient descent.

Initial extensions to full 6 DoF pose optimization focus on handling the non-commutative nature of Euler angle parametrizations. Howard et al.~\cite{howard04} bypass the problem by assuming that roll and pitch are accurately measured, while N\"uchter et al.~\cite{nuechter05} assume that these angles are affected by only small errors. Triebel et al.~\cite{triebel06} in turn parametrize the environment as an ensemble of multiple horizontal surfaces located at different heights. The most notable seminal contributions in 6 DoF pose graph optimization have been made by Grisetti et al.~\cite{grisetti07a,grisetti07b,grisetti09}, who extend the efficient tree-based stochastic optimization method of Olson et al.~\cite{olson06} and further push computational efficiency. Grisetti et al.~\cite{grisetti10} later on present yet another seminal extension of the technique, which enables hierarchical pose graph optimization. Around the same time, Koichi et al.~\cite{koichi10} present the addition of unary factors to include GPS constraints into the optimization. S\"underhauf and Protzel~\cite{suenderhauf12} and Wang and Olson~\cite{wang14} contribute towards the robustness of pose graph optimization. A comprehensive tutorial on pose graph optimization is presented by Grisetti et al.~\cite{grisetti10}, and a popular software framework containing many of the advancements listed here is given by \textit{g2o}~\cite{kuemmerle11}.

It is worth noting that both batch least-squares optimization methods~\cite{triggs00} as well tree-based estimation techniques~\cite{dellaert05,kaess08} have also been introduced for joint optimization over poses and structure. Furthermore, the topics of graph-based optimization have also been investigated in the computer vision community (e.g. robust rotation averaging~\cite{hartley12,govindu13,eriksson18}, robust translation averaging~\cite{wilson14}, and pose synchronization~\cite{crandall11,cui15}).

Our work is a direct extension of the work by Strasdat et al.~\cite{strasdat10}, who explicitly consider the scale invariant nature of single camera geometry. However, in their work they assume that the relative scale between connected nodes in the graph is always a measured, Gaussian-distributed variable. The scale invariant nature is also addressed by Pinies et al.~\cite{pinies08}, who even discuss the partitioning of the graph into several conditionally independent sub-graphs. However, rather than explicitly handling the unknown scale of each sub-graph, they assume that the latter are approximately scaled via the addition of absolute velocity readings during the initialization phase of monocular SLAM. To the best of our knowledge, our work is the first to directly reconcile sub-graphs of arbitrary relative scale through a joint, hybrid, scale jump-aware graph optimization framework.

\section{THEORY}
\label{sec:theory}

We will start by reviewing the traditional formulation of pose graph optimization, its solution, as well its scale-drift-aware adaptation using similarity transformations. Next, we will present our modified hybrid pose graph formulation which explicitly takes into account the existence of unknown scale jumps. To conclude, we will present a discussion on degenerate cases and a method to identify whether or not a single, globally consistent scale factor can be reconciled.

\subsection{Brief review of scale drift-aware pose graph optimization}

The problem of pose graph optimization can be abstracted as follows. Let $\tilde{\mathbf{T}}_{ij}$ be the measured relative transformation between two nearby sensor frames $i$ and $j$. It is a Euclidean transformation and can be used to linearly map points in homogeneous representation from sensor frame $j$ to $i$, hence
\begin{eqnarray}
  \left[ \begin{matrix} ^i\mathbf{p}^T & 1 \end{matrix}\right]^T & = & \tilde{\mathbf{T}}_{ij} \left[ \begin{matrix} ^j\mathbf{p}^T & 1 \end{matrix}\right]^T \\
  & = & \left[ \begin{matrix} \tilde{\mathbf{R}}_{ij} & \tilde{\mathbf{t}}_{ij} \\ \mathbf{0} & 1 \end{matrix}\right] \left[ \begin{matrix} ^j\mathbf{p}^T & 1 \end{matrix}\right]^T. \nonumber
\end{eqnarray}
The estimated variables in pose graph optimization are given by the absolute sensor frame poses $\mathbf{T}_i$ which are defined such that they linearly map points from a sensor frame to a global reference frame, that is
\begin{eqnarray}
  \left[ \begin{matrix} ^\mathcal{W}\mathbf{p}^T & 1 \end{matrix}\right]^T & = & \mathbf{T}_{i}  \left[ \begin{matrix} ^i\mathbf{p}^T & 1 \end{matrix}\right]^T \\
  & = & \left[ \begin{matrix} \mathbf{R}_{i} & \mathbf{t}_{i} \\ \mathbf{0} & 1 \end{matrix}\right] \left[ \begin{matrix} ^i\mathbf{p}^T & 1 \end{matrix}\right]^T. \nonumber
\end{eqnarray}
The goal of pose graph optimization consists of estimating absolute sensor frame poses $\mathbf{T}_i$ such that their total discrepancy with respect to each measured relative pose is minimized. Formally, the objective is given by
\begin{equation}
  \underset{\boldsymbol{\theta}_1, \ldots, \boldsymbol{\theta}_N}{\operatorname{argmin}} \sum_{k=1}^{M} \left\| \operatorname{t2v}
  \left(\left(\mathbf{T}\left(\boldsymbol{\theta}_{j_k}\right)\right)^{-1}\mathbf{T}\left(\boldsymbol{\theta}_{i_k}\right)\tilde{\mathbf{T}}_{i_kj_k} \right)\right\|^2,
\end{equation}
which minimizes the sum of squared deviations from identity transformation of the concatenations of each measured relative pose $\tilde{\mathbf{T}}_{i_kj_k}, k=1,\ldots,M$ and the two corresponding, optimized absolute poses. The latter are represented minimally by the 6-vectors $\boldsymbol{\theta}=\left[\begin{matrix}\mathbf{t}^T & \boldsymbol{\phi}^T\end{matrix}\right]^T$ containing the translation $\mathbf{t}$ and the Rodriguez vector $\boldsymbol{\phi}$. $\mathbf{T(\boldsymbol{\theta})}$ uses the Riemannian exponential map to obtain the corresponding rotation matrix, and $\operatorname{t2v}(\mathbf{T})$ makes use of the Riemannian logarithmic map to again go back to minimal representation. It is intuitively clear that when the absolute poses are consistent with the relative pose measurements they will form transformation cycles and the corresponding minimal vector will be zero. Any inconsistencies will result in a non-zero vector, and thus contribute to the overall energy. In practice, we will often minimize the robust, covariance-reweighted alternative
\begin{equation}
  \underset{\boldsymbol{\theta}_1, \ldots, \boldsymbol{\theta}_N}{\operatorname{argmin}} \sum_{k=1}^{M} \rho\left( \mathbf{r}_1(\boldsymbol{\theta}_{i_k},\boldsymbol{\theta}_{j_k})^T \boldsymbol{\tilde{\Omega}}_{i_k,j_k} \mathbf{r}_1(\boldsymbol{\theta}_{i_k},\boldsymbol{\theta}_{j_k}) \right),
\end{equation}
where $\mathbf{r}_1(\boldsymbol{\theta}_{i_k},\boldsymbol{\theta}_{j_k})=\operatorname{t2v}
  \left(\left(\mathbf{T}\left(\boldsymbol{\theta}_{i_k}\right)\right)^{-1} \mathbf{T}\left(\boldsymbol{\theta}_{j_k}\right)\right) - \tilde{\boldsymbol{\theta}}_{i_kj_k}$ is the relative pose consistency term, $\tilde{\boldsymbol{\theta}}_{i_kj_k}$ is the minimal representation of the measured relative pose, $\tilde{\boldsymbol{\Omega}}_{i_k,j_k}$ the corresponding information matrix, and $\rho(\cdot)$ a robust cost function (e.g. Huber-loss) that will down-weight the influence of outliers. In case of global estimation, the iterative updates are finally calculated using sparse Cholesky decomposition.

In the monocular case, the initial relative pose estimate is given from direct frame-to-frame relative pose estimation, which is scale invariant. In sub-sequent frames, the algorithm then performs local map tracking, thus aiming at scale propagation. In an ideal scenario, the estimated relative poses therefore all differ from reality by 1) the presence of noise, and 2) a globally-consistent scale factor. We could simply apply the above pose-graph optimization formulation, which would find absolute sensor frame poses that would all be affected by the same global scale parameter. However, just like other parts of the algorithm, the scale propagation mechanism is affected by measurement uncertainties, thus causing the scale to drift over time and indeed turn into a local variable. For this reason, a naive application of the above pose graph optimization objective will lead to sub-optimal results.

In order to take into account the local nature of the scale factor, Strasdat et al.~\cite{strasdat10} have proposed scale-aware pose-graph optimization based on similarity transformations. The absolute pose is thereby represented as
\begin{equation}
\mathbf{T}_i = \left[\begin{matrix} e^{s_i} \mathbf{R}_i & \mathbf{t}_i \\ \mathbf{0} & 1 \end{matrix}\right],
\end{equation}
and the minimal vectors used inside the pose graph optimizer are now given by the 7-vectors $\boldsymbol{\theta} = \left[\begin{matrix}\mathbf{t}^T & \boldsymbol{\phi}^T & s\end{matrix}\right]^T$. The estimated relative transformations become
\begin{eqnarray}
& & \left(\mathbf{T}\left(\boldsymbol{\theta}_{i_k}\right)\right)^{-1} \mathbf{T}\left(\boldsymbol{\theta}_{j_k}\right) = \\
& & \left[\begin{matrix}  e^{s_j-s_i} \mathbf{R}(\boldsymbol{\phi}_{i_k})^T\mathbf{R}(\boldsymbol{\phi}_{j_k})  & e^{-s_i} \mathbf{R}(\boldsymbol{\phi}_{i_k})^T (\mathbf{t}_j-\mathbf{t}_i) \\ \mathbf{0} & 1 \end{matrix}\right], \nonumber
\end{eqnarray}
and a few interesting facts can be noted:
\begin{itemize}
\item The hypothesized relative poses are similarity transformations.
\item The scale of neighboring sensor frames is often similar. As required, the scale of the hypothesized relative pose becomes 1 in that situation ($s_i\approx s_j \Rightarrow e^{s_j-s_i}\approx 1$).
\item In a loop closure situation, the scale difference could potentially be very large. Again, this is properly reflected ($s_i\neq s_j \Rightarrow e^{s_j-s_i}\neq 1$).
\item The optimized relative translation is properly scaled in order to enable comparison against the possible scaled relative translation measurements.
\end{itemize}

\subsection{Scale-jump aware hybrid pose graph optimization}

\begin{figure*}[t]
    \centering
    \includegraphics[width=\textwidth]{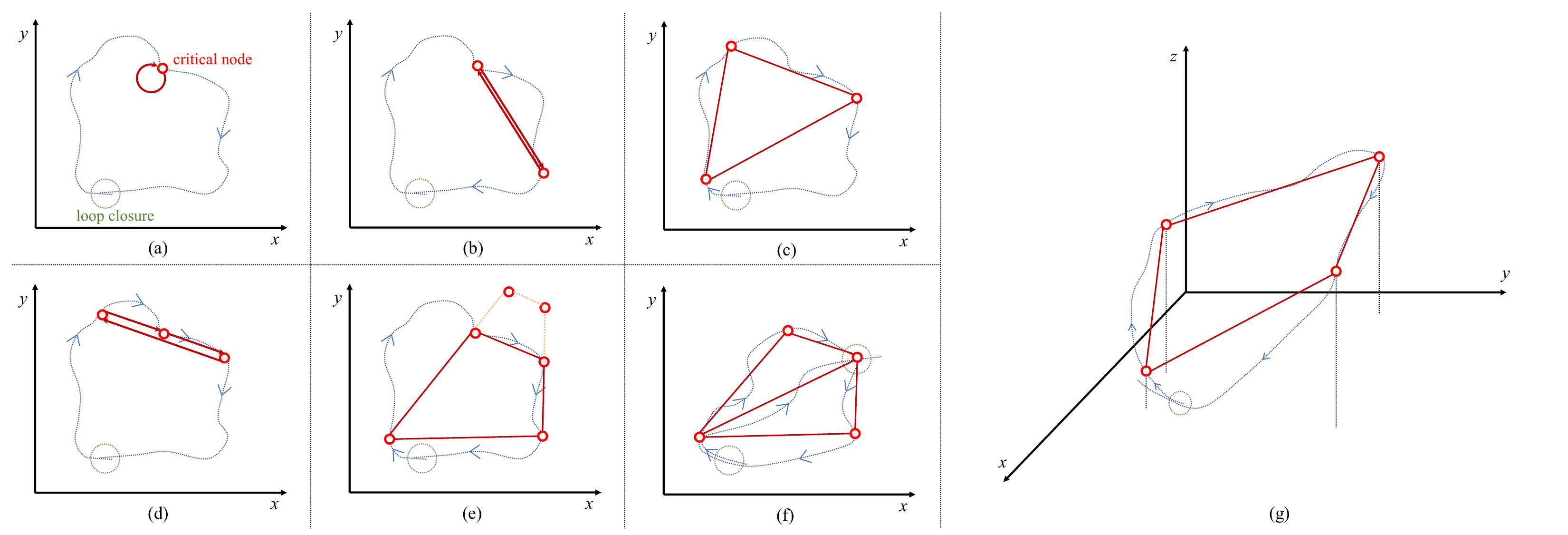}
    \caption{Visualization of possible loopy graphs with tracking failures (i.e. critical nodes). (a), (b) and (c) show loops along which only one, two, or three tracking failures occur. (d) illustrates the special case in which the Euclidean position of the three critical nodes are aligned. (e) shows a planar scenario with 4 critical nodes. (f) illustrates how additional bars can reinstall global scale consistency. (g) indicates a 3D arrangement of 4 non-planar critical nodes. All configurations except (d) and (e) lead to global scale consistency. For (d) and (e), an-isotropic scaling is possible. See text for details.}
    \label{fig:cases}
\end{figure*}

While adding the scale as an optimization variable helps to account for scale variations, its potential is nonetheless limited to situations in which the measured relative scale is known up to noise. During regular tracking, the relative scale is simply assumed to be 1, and hence the formulation only supports the estimation of relatively slow scale changes over time (i.e. drift). During loop closure, the relative scale is calculated by for example generalized Procrustes alignment. In both cases, the relative scale is represented as a Gaussian variable.

In practice, we may easily face situations in which the relative scale factor can no longer be reflected by a Gaussian estimate with mean and standard deviation. A simple example is given when the algorithm loses the local map and needs to reinitialize. The scale after initialization is in principle arbitrary, and hence the relative scale becomes a uniformly distributed variable. In order to take into account the possible existence of edges with unknown relative pose, we propose an alternative Hybrid Pose Graph Optimization (HPGO) that makes use of the modified residual term
\begin{eqnarray}
& & \mathbf{r}_2(\boldsymbol{\theta}_{i_k},\boldsymbol{\theta}_{j_k})= \\ 
& & \left[ \operatorname{t2v}
  \left(\left(\mathbf{T}\left(\boldsymbol{\theta}_{i_k}\right)\right)^{-1} \mathbf{T}\left(\boldsymbol{\theta}_{j_k}\right)\right) - \tilde{\boldsymbol{\theta}}_{i_kj_k}\right] \left[ \begin{matrix} \mathbf{I}_{6\times 6} & \mathbf{0} \\ \mathbf{0}^T & 0 \end{matrix} \right] \nonumber
\end{eqnarray}
along edges where the relative scale factor is unknown. Note that, rather than introducing an alternative noise model, the term simply ignores errors in the relative scale, thus permitting arbitrary scale changes to occur. The complete objective of HPGO is finally given by
\begin{equation}
  \underset{\boldsymbol{\theta}_1, \ldots, \boldsymbol{\theta}_N}{\operatorname{argmin}} \sum_{k=1}^{M} \rho\left( \mathbf{r}_{d_k}(\boldsymbol{\theta}_{i_k},\boldsymbol{\theta}_{j_k})^T \boldsymbol{\tilde{\Omega}}_{i_k,j_k} \mathbf{r}_{d_k}(\boldsymbol{\theta}_{i_k},\boldsymbol{\theta}_{j_k}) \right),
\end{equation}
where
\begin{itemize}
  \item $d_k = 1$ if the $k$-th edge is a regular edge along which a Gaussian estimate for the relative scale exists, and
  \item $d_k = 2$ otherwise.
\end{itemize}

Note that there may well be situations in which the relative scale is unknown while other parameters are still given. Let us consider the example of a platform that is able to execute purely rotational displacements (e.g. a robotic vacuum cleaner). It is easily possible that the local map is pushed out of the current field of view. While the relative transformation (i.e. rotation) remains measurable, the eventual re-initialization will possibly lead to a jump in the local scale factor. Interestingly, if a loop closure occurs, it is still possible to reconcile a globally consistent graph for which the individual segments after each algorithm re-initialization remain consistently scaled. In the following, we will analyze the situations in which a globally consistent trajectory can still be recovered.

\subsection{Feasibility study}

In order to explain why and when a globally consistent scale can be reconciled, let us consider a geometric graph abstraction. We identify the nodes which are located at the end of edges for which the relative scale is unknown, and denote them \textit{critical nodes}. We record the true location of all critical nodes in a Euclidean space, and create pairwise connections between nodes if they are directly connected along the actual trajectory (i.e. without passing through other critical nodes). The obtained geometric graph abstraction is a construct of bars each one representing one part of the entire trajectory that---owing to known relative scale measurements---is consistently scaled after optimization.

The above question may now be addressed as follows: \textit{Besides a trivial globally consistent scaling of the entire bar construct, are there any subsets of bars that can be scaled individually without requiring a change of the relative angles between the bars?} The requirement that the relative orientation of the bars needs to remain unchanged stems from the fact that indeed the orientation of the platform at the end of a trajectory segment must remain consistent with the orientation at the beginning of the next segment.

Figure~\ref{fig:cases} illustrates a few basic cases that will serve to study the feasibility of global scale reconciliation.  First, it is intuitively clear that a globally consistent scale can only be obtained if the graph contains at least one loop. For critical nodes that are not part of a loop, some trajectory segments will retain undetermined scale. Loopy cases are as follows:
\begin{itemize}
    \item Figure~\ref{fig:cases} (a): If there is only one critical node in the loop, the bar construct will degenerate. Traveling along the loop will always return to the same critical node, hence the latter is connected to itself by a bar with zero length. As a result, there is only one scale factor, and global consistency is always achievable.
    \item Figure~\ref{fig:cases} (b): If there are two nodes in the graph abstraction, we would have two possible trajectory segments connecting the two nodes, and the respective bars would simply be the same. As they would always have equal scale, a loop with two tracking failures would always lead to reconcilable global consistency.
    \item Figure~\ref{fig:cases} (c): For three critical nodes, we generally obtain a triangular construct of bars. It is clear that the requirement of unchanged angles means that we would require similar triangles, and all bars would again have to be isotropically scaled. A degenerate case is given if the triangle collapses and all bars become parallel (cf. Figure~\ref{fig:cases} (d)). In the latter case, it would become possible to apply different scales to the three segments without affecting the angular consistency requirement.
    \item Figure~\ref{fig:cases} (e): For a single loop with four critical nodes, a single scale can no longer be reconciled in the planar case. A simple example is given by a rectangular arrangement of the bars, which would enable independent scaling of parallel bars. A similar situation occurs however for an arbitrary quadrilateral arrangement.
    \item Figure~\ref{fig:cases} (f): Things change as soon as more connections are added to the graph. Compared to the previous case, if a diametrical connection is added, an-isotropic scaling would again violate the angle preservation requirement. A general rule can be stated as follows: If every critical node is part of at least one non-degenerate triangle, global scale consistency can be reconciled.
    \item Figure~\ref{fig:cases} (g): It is important to note that all previous scenarios only consider the planar case. In the 3D case, for a non-planar arrangement of 4 nodes, global scale consistency can always be achieved. However, if there is at least one loop with more than four critical nodes, globally consistent scale can no longer be reconciled (irrespectively of the dimensionality of the problem).
\end{itemize}

From a mathematical perspective, the possibility of a global scale reconciliation may be analyzed as follows. Let there be $B$ bars in the graph, and let $\mathbf{p}_1,\ldots,\mathbf{p}_N$ be the absolute positions of the critical nodes. Let the starting and ending nodes of the $k$-th bar be given by $\{\mathbf{p}_{s_k},\mathbf{p}_{e_k}\}$. If no relative angles between bars are allowed to change, it indeed means that the absolute orientation of each bar needs to remain unchanged. Let $\mathbf{v}_1,\ldots,\mathbf{v}_B$ be vectors representing the original globally registered bars. In order for the angular constraint to be satisfied, we need the end points of each bar to remain a scaled versions of the original bar vector, i.e.
\begin{equation}
    (\mathbf{p}_{e_k} - \mathbf{p}_{s_k}) - \lambda_k \mathbf{v}_k = \mathbf{0}.
\end{equation}
Stacking all $B$ constraints in a large matrix, we obtain
\small
\begin{equation}
  \left[ \begin{matrix}
    \mathcal{I}_{3\times 3N}(s_1,e_1) & \mathbf{v}_1 &       &       &       &              \\
    \cdot                             &              & \cdot &       &       &              \\
    \cdot                             &              &       & \cdot &       &              \\
    \cdot                             &              &       &       & \cdot &              \\
    \mathcal{I}_{3\times 3N}(s_B,e_B) &              &       &       &       & \mathbf{v}_B \\
    \left[ \mathbf{I}_{3\times 3} \text{ } \mathbf{0}_{3\times 3(N-1)} \right] & \mathbf{0} & \cdot & \cdot & \cdot & \mathbf{0}
  \end{matrix} \right]
  \left[ \begin{matrix} \mathbf{p}_1 \\ \cdot \\ \cdot \\ \mathbf{p}_N \\ \lambda_1 \\ \cdot \\ \cdot \\ \lambda_B \end{matrix}  \right] = \mathbf{A} \mathbf{x} = \mathbf{0},
  \label{equ:evaluation_matrix}
\end{equation}
\normalsize
where $\mathcal{I}_{3\times 3N}(s_1,e_1)$ is a horizontal concatenation of $N$ matrices that are all $3 \times 3$ zero matrices except for the $e_k$-th matrix which is $\mathbf{I}$ and the $s_k$-th matrix which is $-\mathbf{I}$. Note that the last row is added in order to remove positional gauge freedom. The requirement of at most a single, global scale consistency can now easily be analysed by checking the rank of $\mathbf{A}$. If $\mathbf{A}$ has a single null-space vector, a globally consistent scale can be reconciled. However, if $\mathbf{A}$ has a rank deficiency larger than 1, a global scale may no longer be reconciled. Note that in practice, rotational drift will make it impossible to figure out globally oriented bars before the actual pose graph optimization has finished, hence the overall graph consistency can only be analyzed retrospectively.

\section{EXPERIMENTS}

\begin{figure*}[t]
  \centering
  \begin{minipage}[t]{0.49\textwidth}
    \centering
    \subfigure[Triangle]{
    \includegraphics[width=0.32\textwidth]{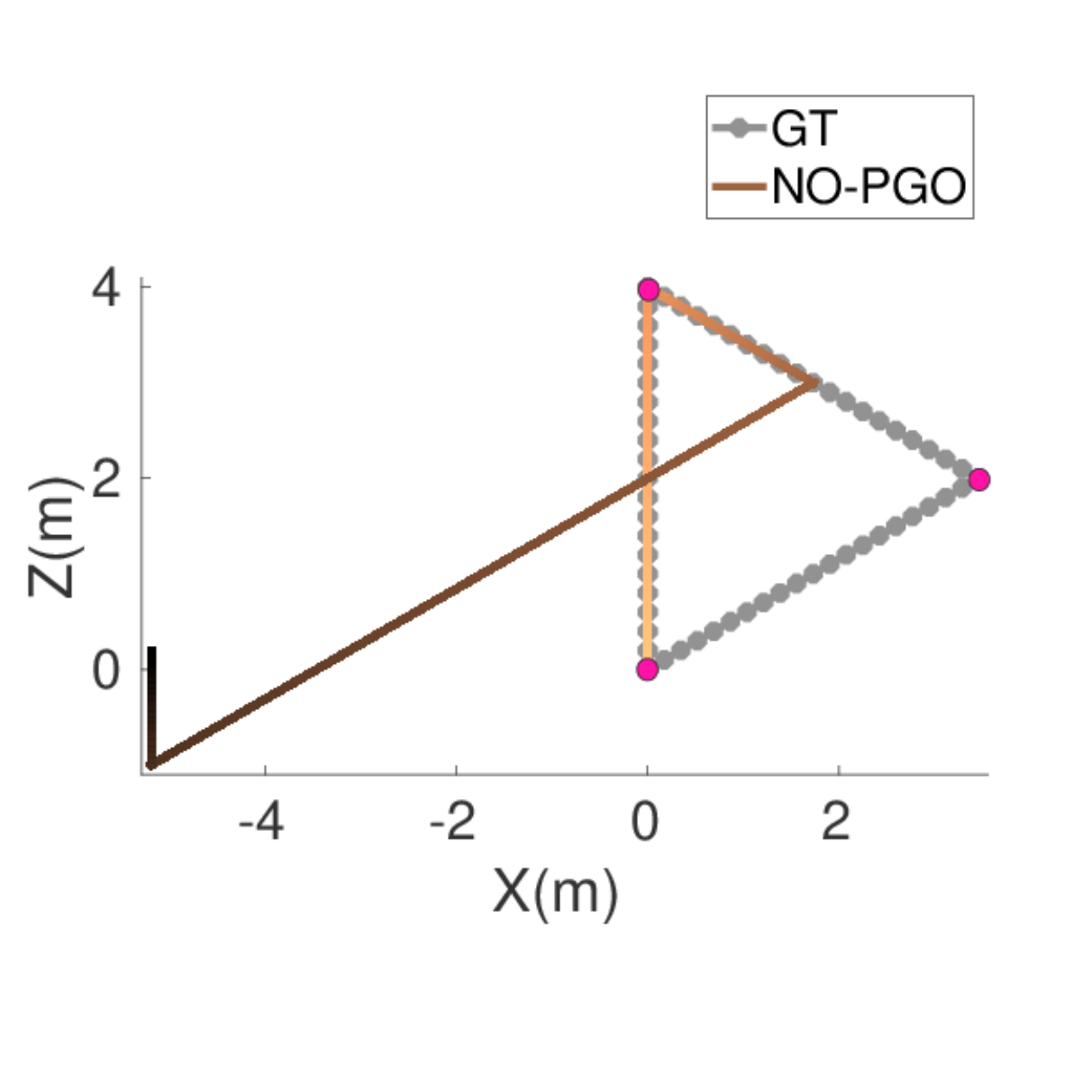}
    \includegraphics[width=0.32\textwidth]{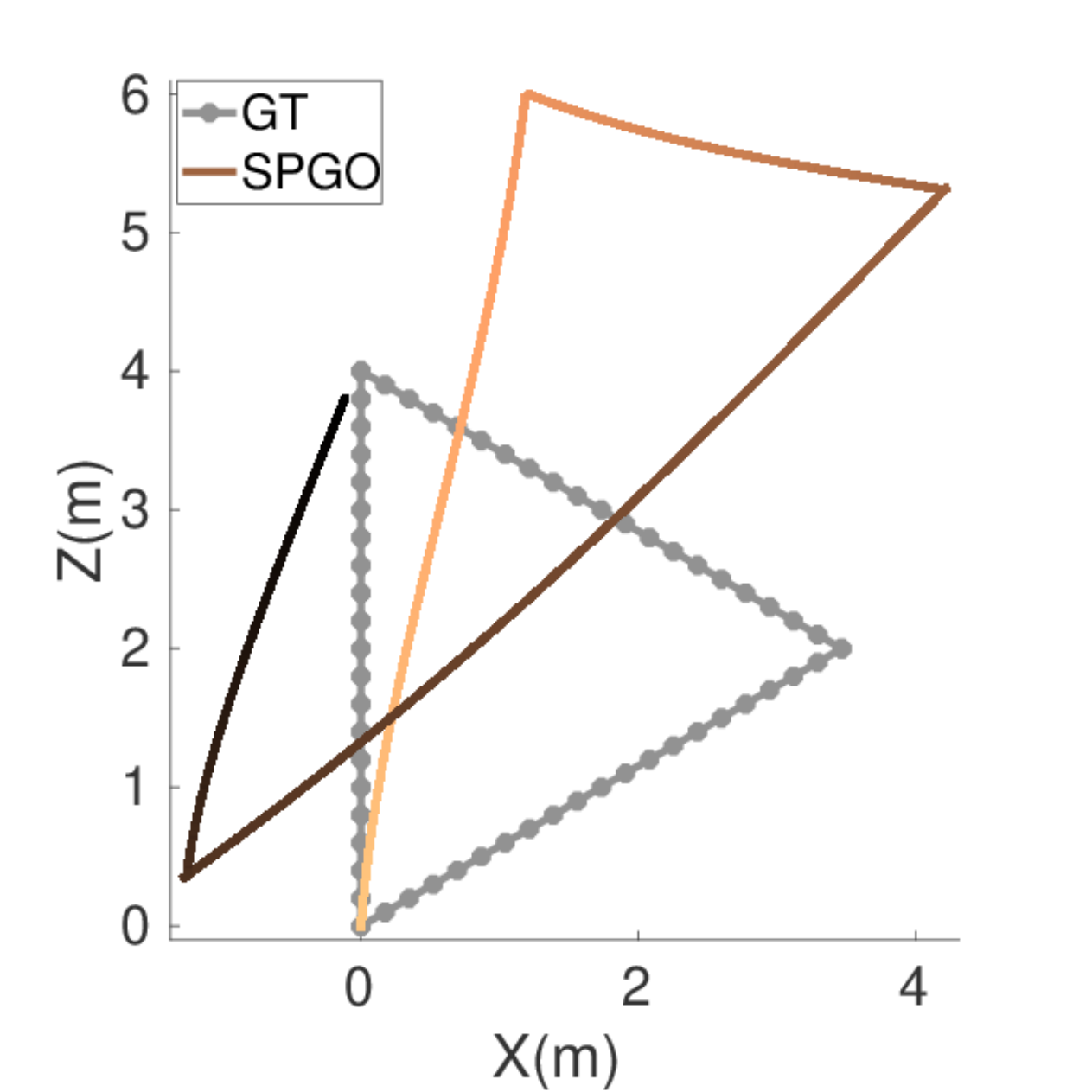}
    \includegraphics[width=0.32\textwidth]{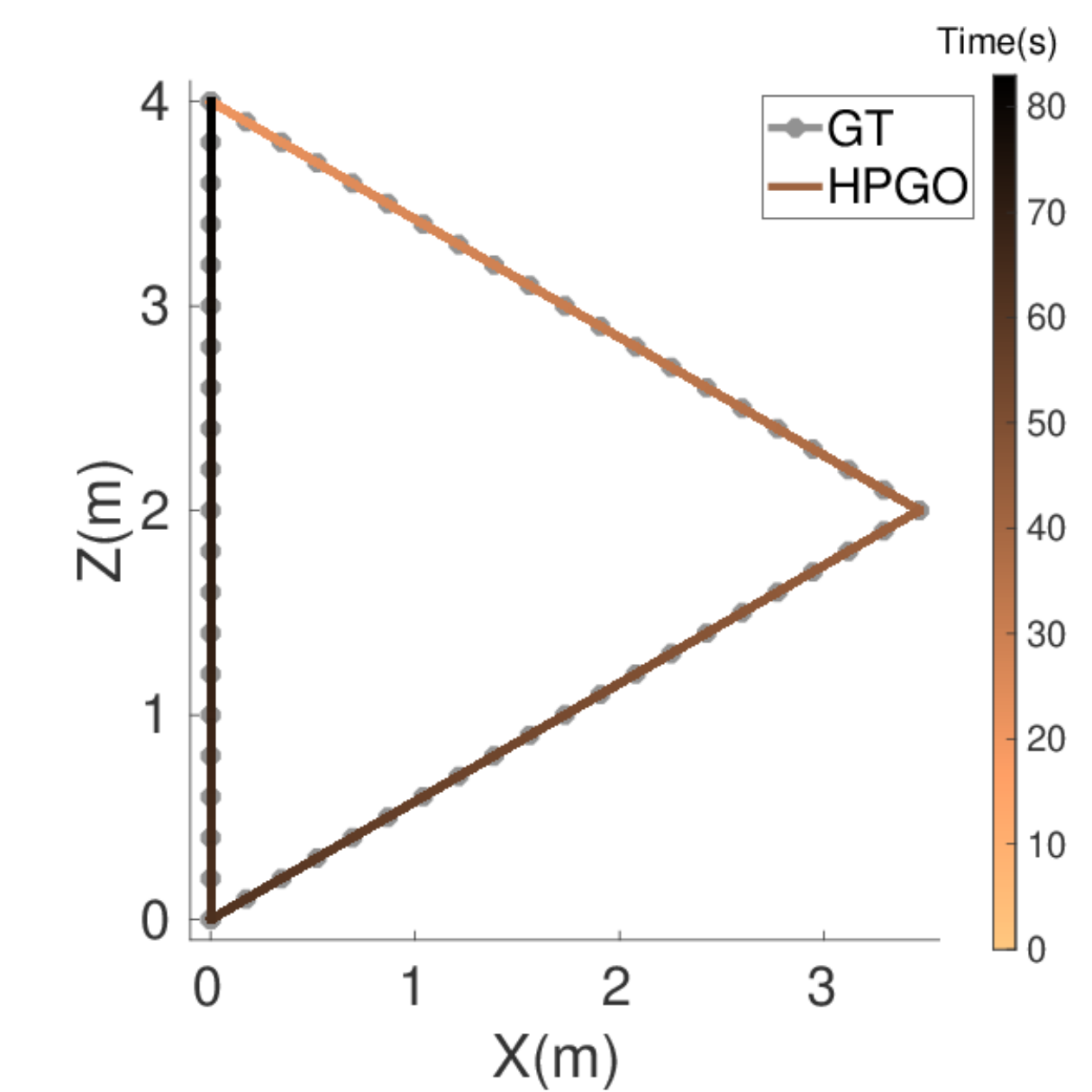}
    }

  \end{minipage}
  \vrule
  \begin{minipage}[t]{0.49\textwidth}
    \centering
    \subfigure[Rectangle]{
    \includegraphics[width=0.32\textwidth]{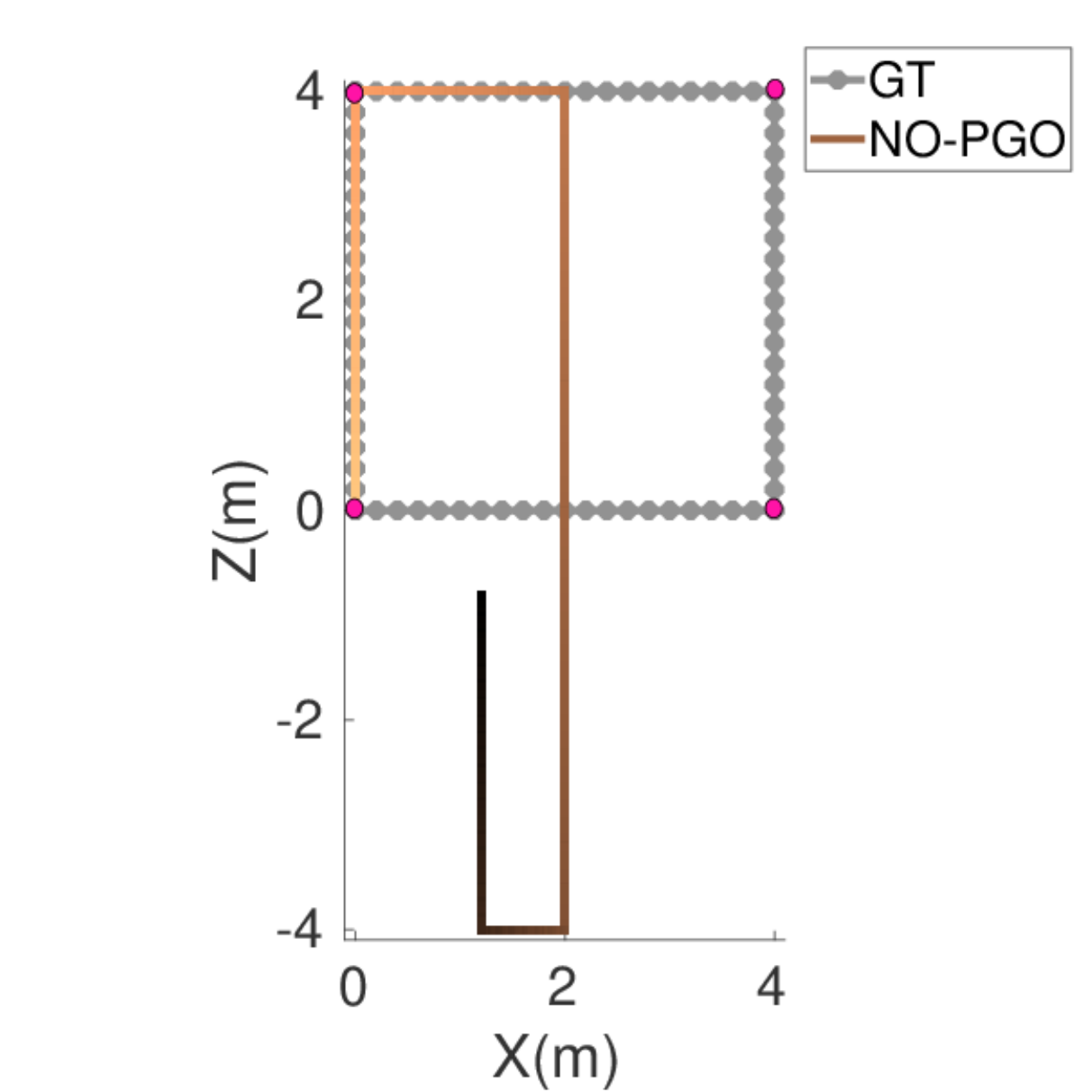}
    \includegraphics[width=0.32\textwidth]{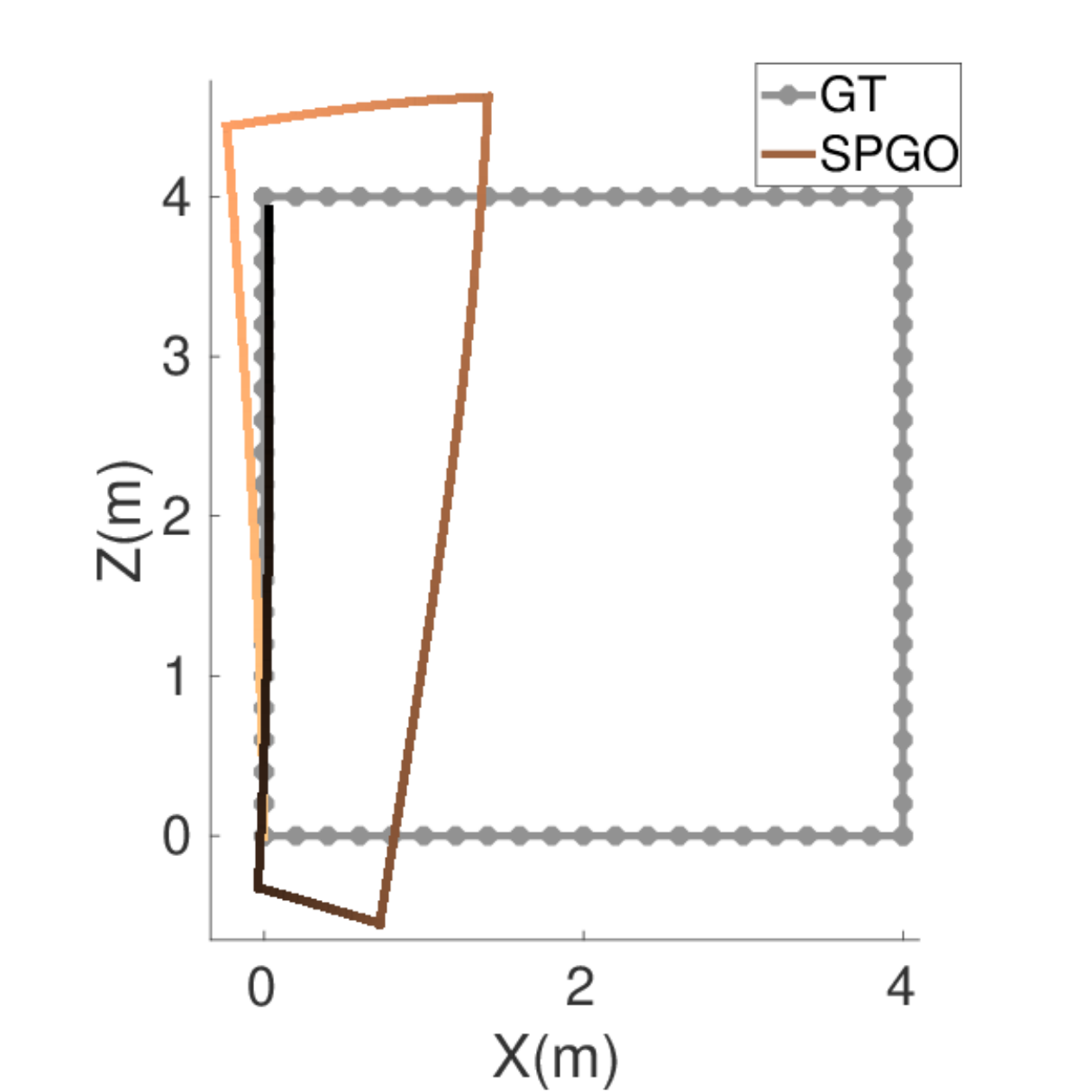}
    \includegraphics[width=0.32\textwidth]{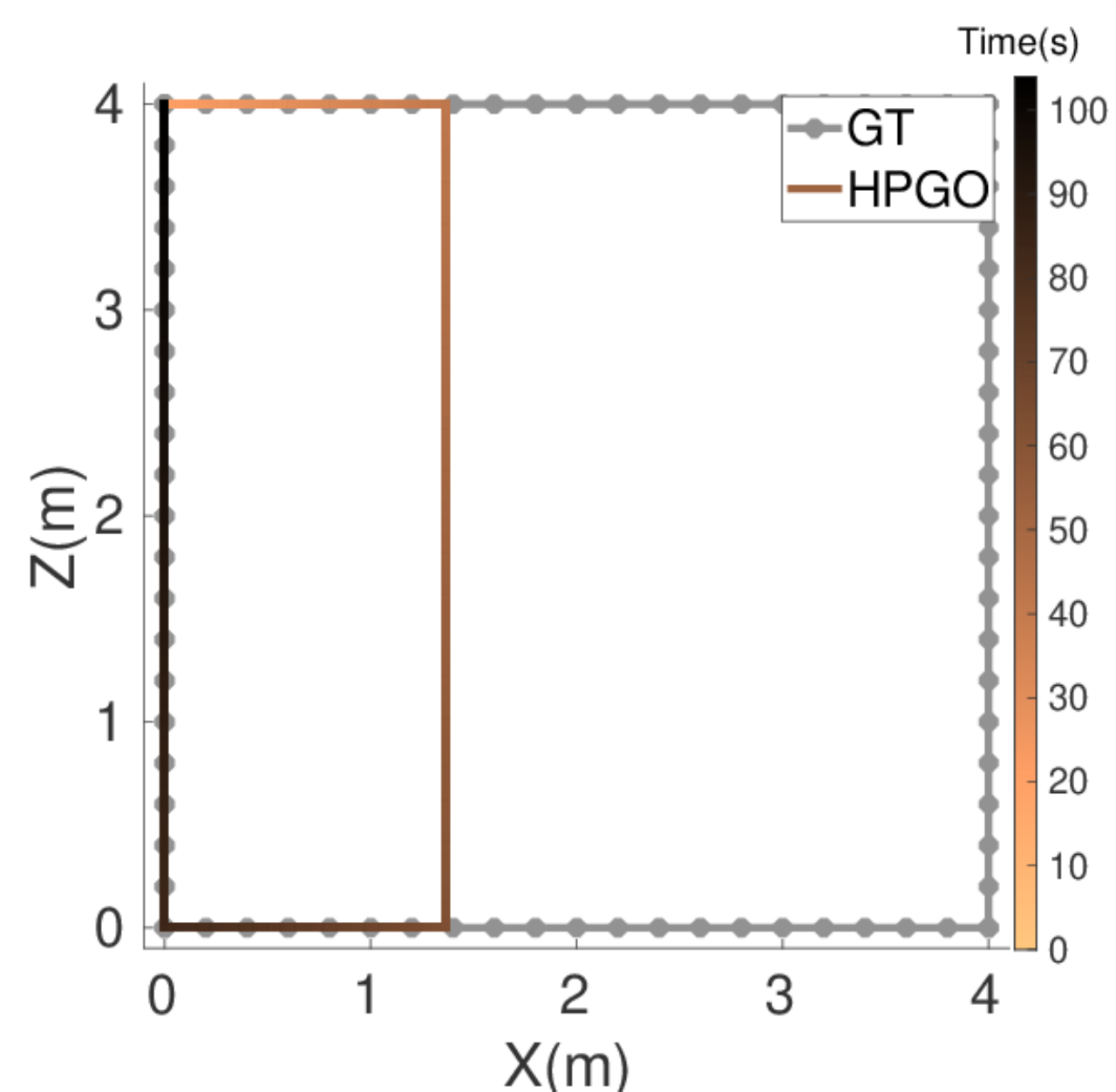}
    }
  \end{minipage}

  \vspace{0.02cm}

  \hrulefill
  
  \vspace{0.02cm}
  
  \begin{minipage}[t]{0.49\textwidth}
    \centering
    \subfigure[Circle with 4 segments]{
    \includegraphics[width=0.32\textwidth]{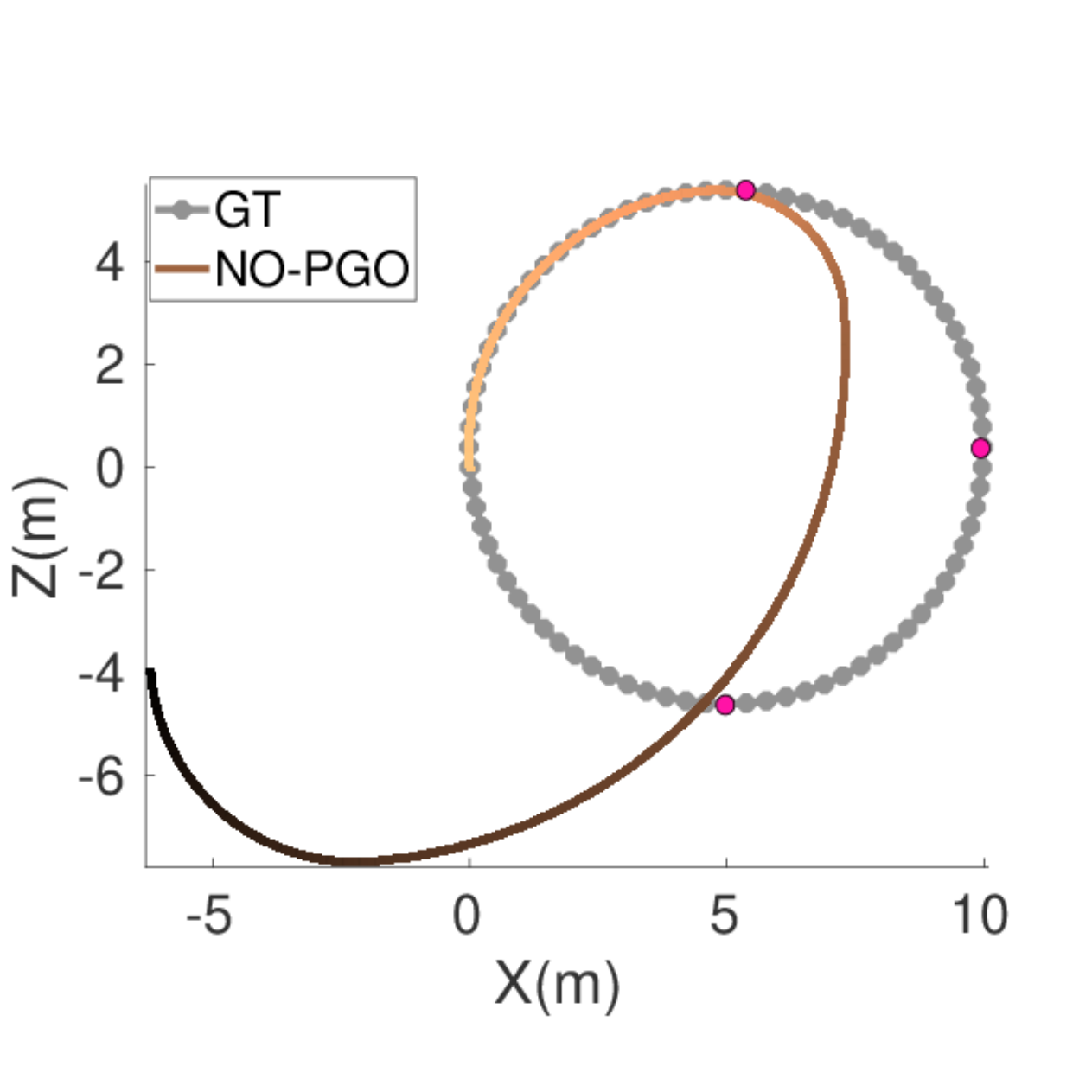}
    \includegraphics[width=0.32\textwidth]{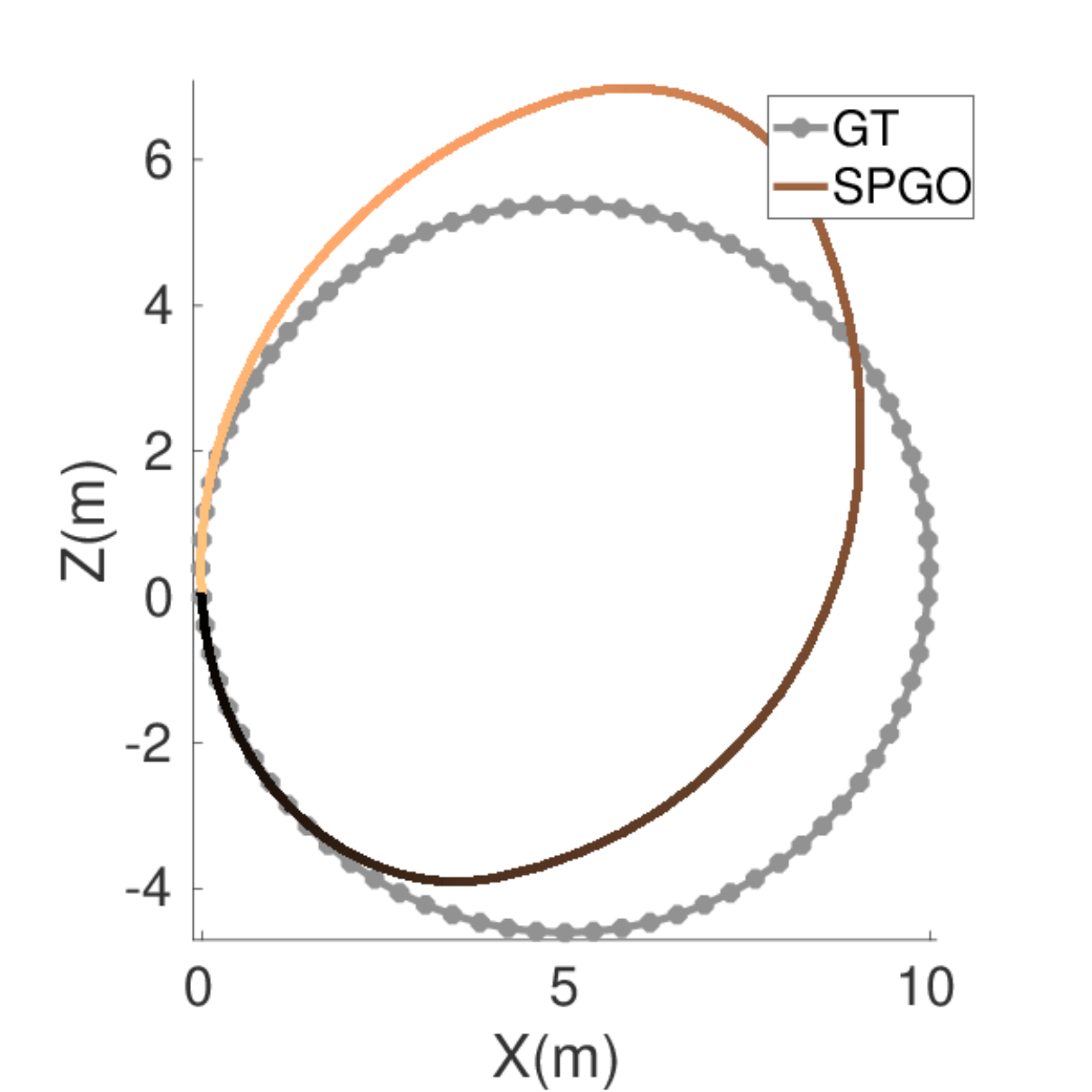}
    \includegraphics[width=0.32\textwidth]{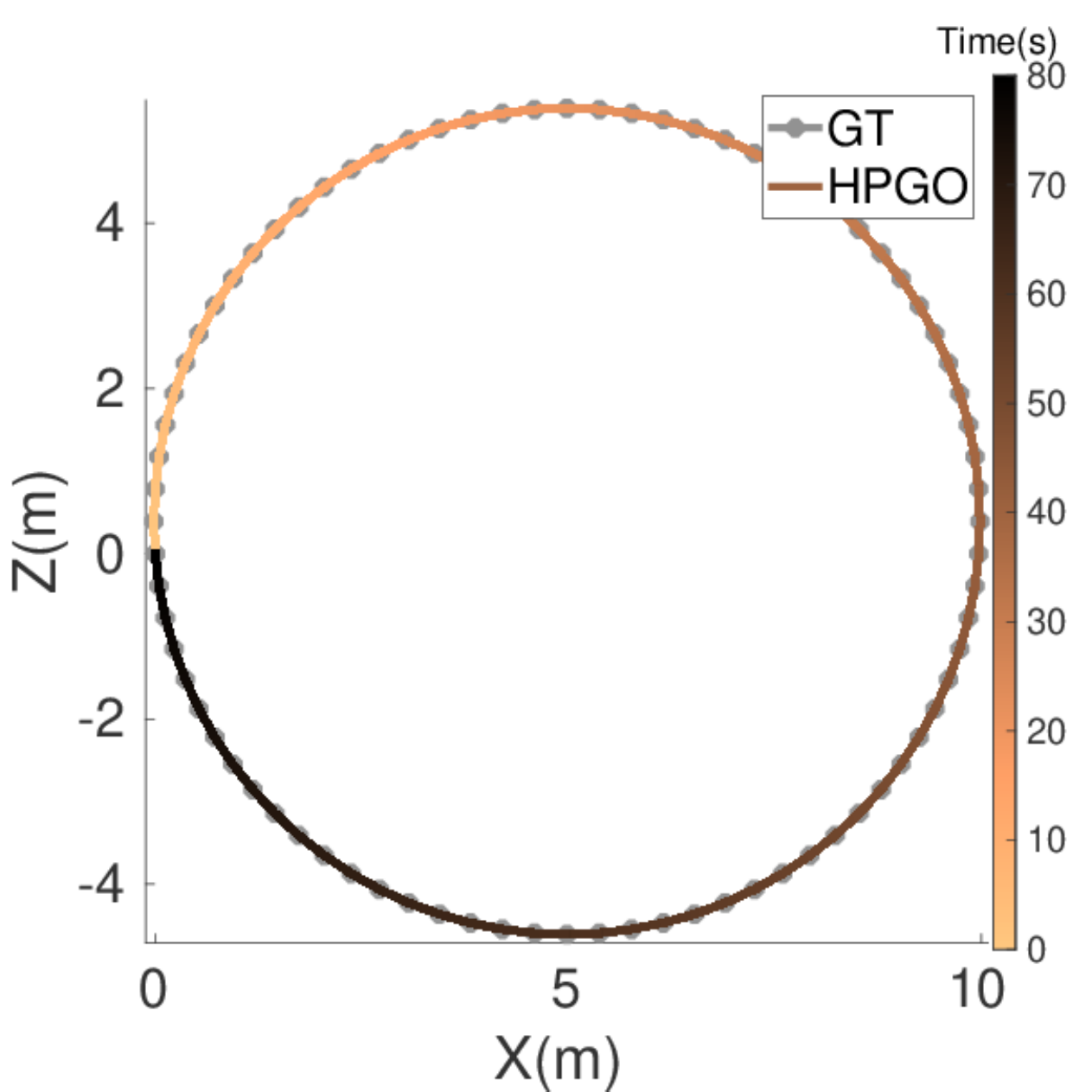}
    }
  \end{minipage}
  \vrule
  \begin{minipage}[t]{0.49\textwidth}
    \centering
    \subfigure[Circle with 5 segments]{
    \includegraphics[width=0.32\textwidth]{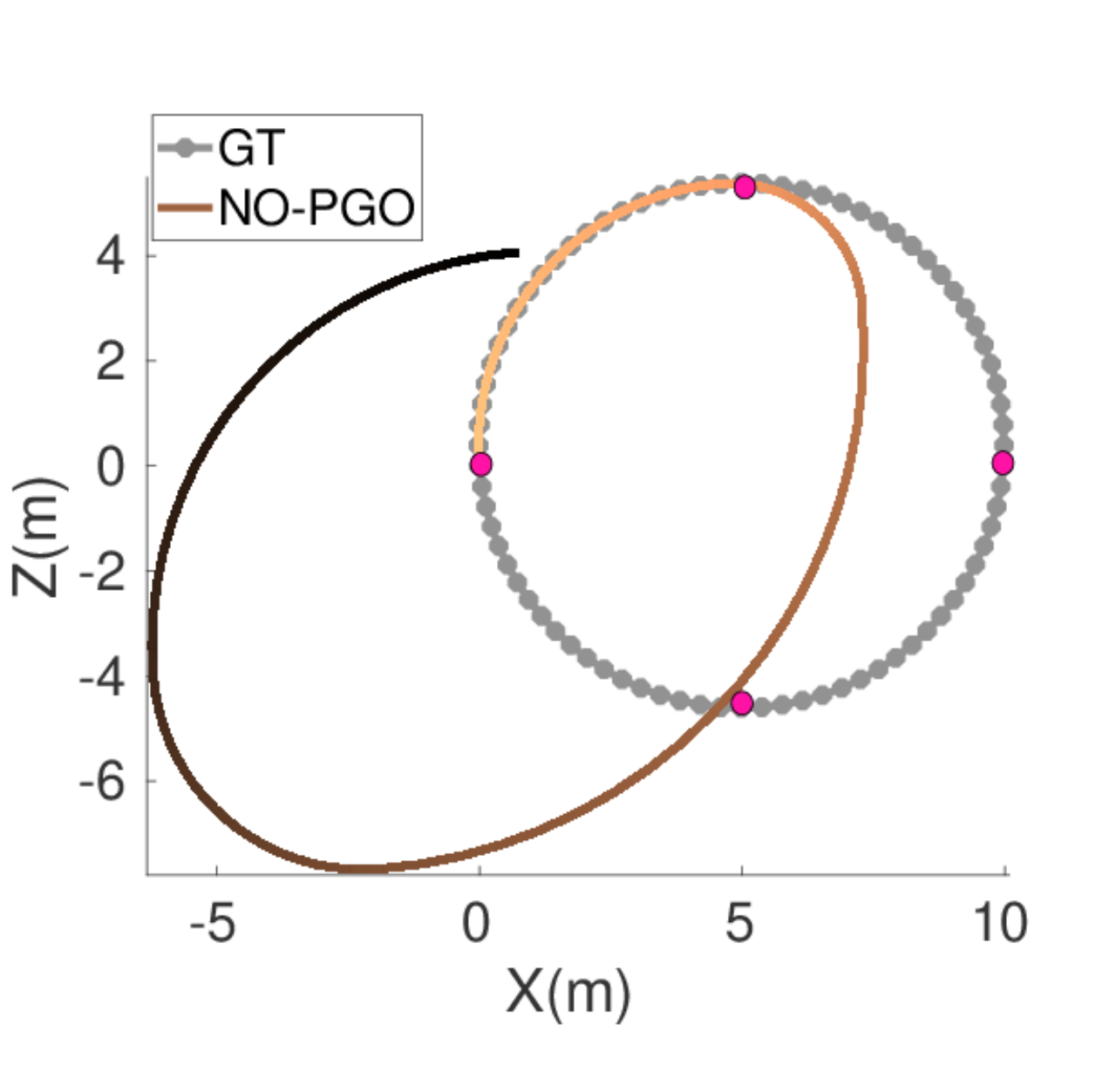}
    \includegraphics[width=0.32\textwidth]{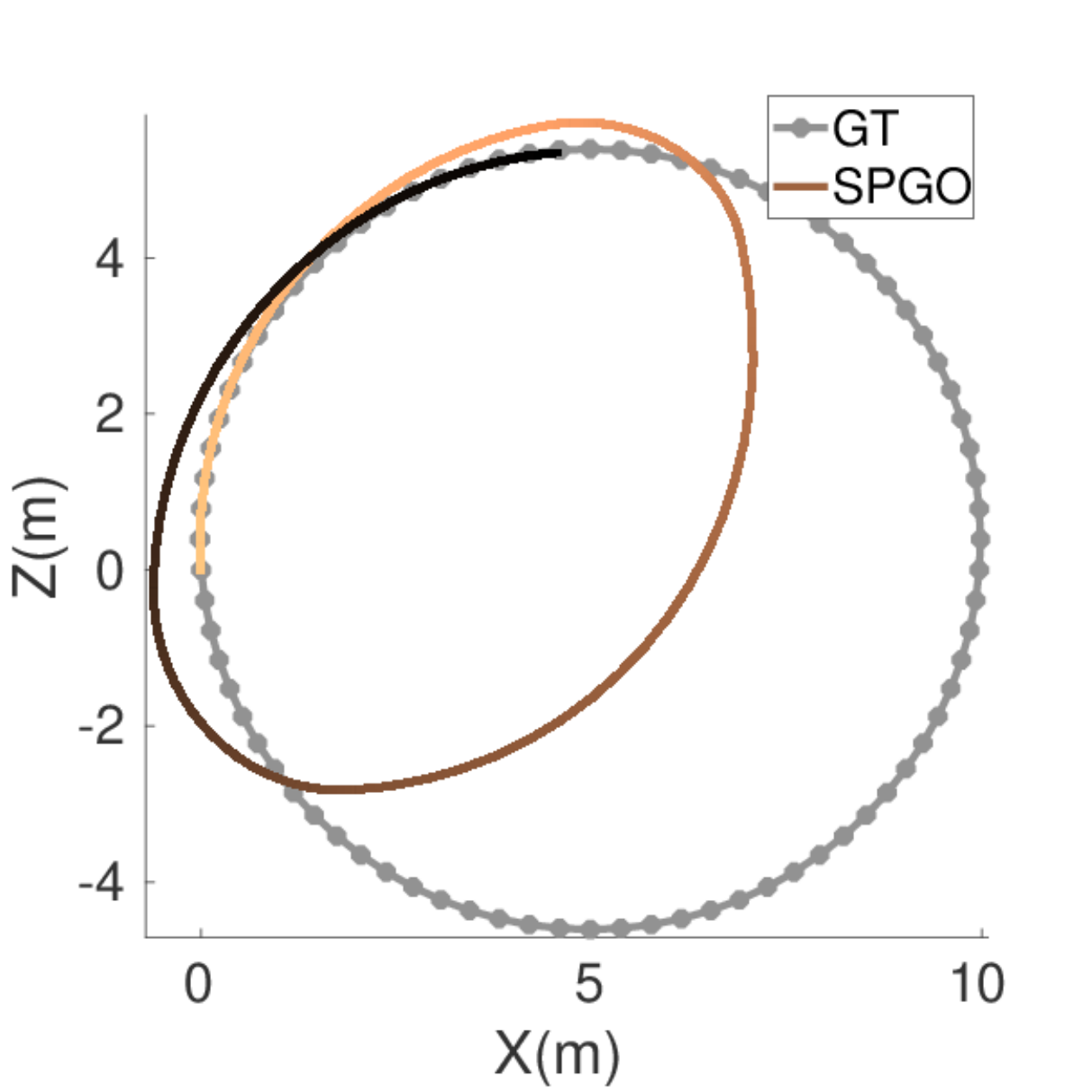}
    \includegraphics[width=0.32\textwidth]{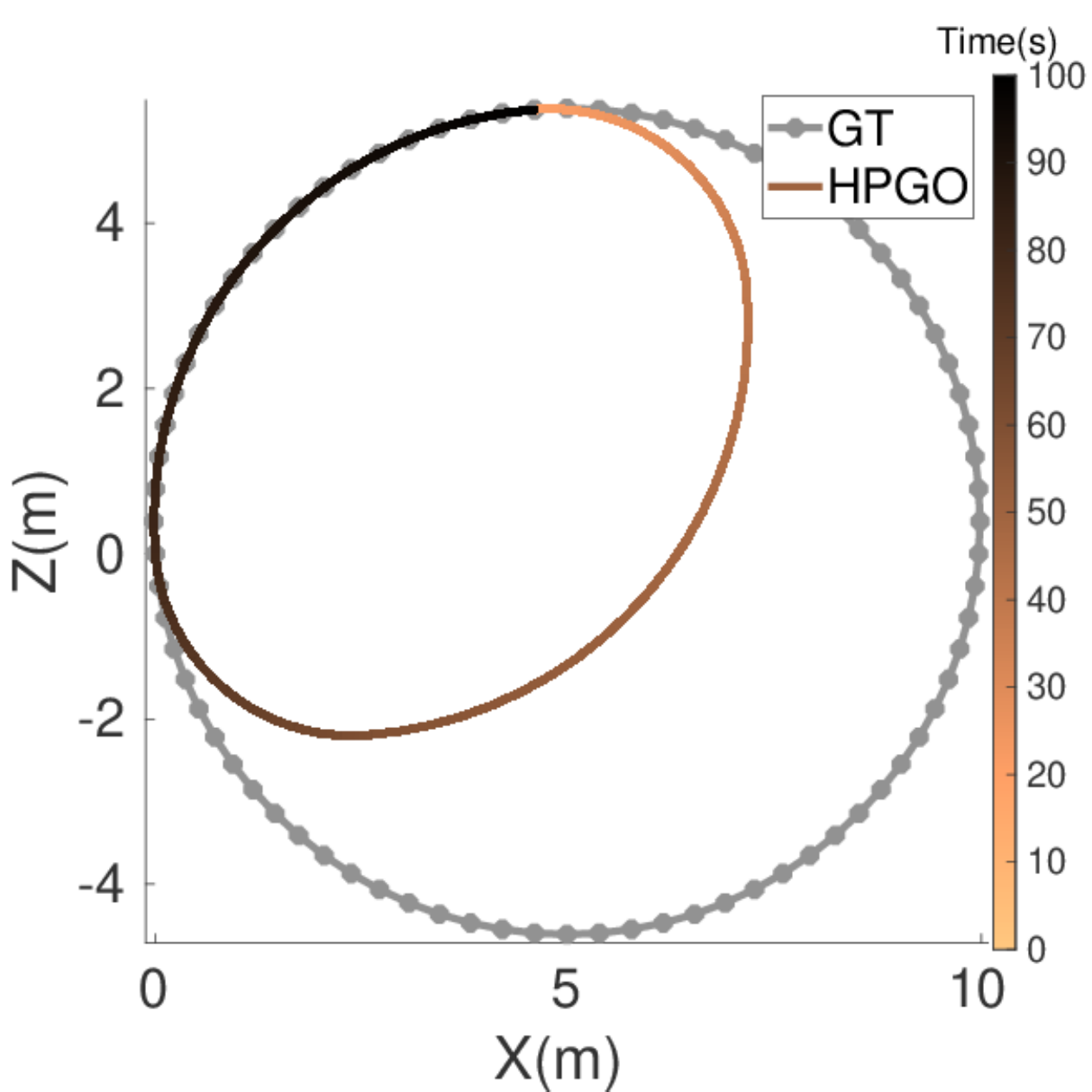}
    }
  \end{minipage}
  
  \caption{Simulation results obtained for differently shaped planar trajectories. From left to right, each triplet indicates the original un-optimized trajectory, the result obtained by scale-drift aware pose graph optimization (SPGO), and the result obtained by our hybrid optimization framework (HPGO). The color reflects the time of each frame (light colors first, dark colors last). Critical nodes are indicated in pink.}
  \label{fig:simulation_traj}
\end{figure*}
\label{sec:experiments}

We now proceed to the experimental evaluation of our novel hybrid pose graph optimization and demonstrate the validity of our findings on both simulated and real data. In both cases, the un-optimized input to the optimizer is given in the form of concatenated trajectory segments that exhibit scale jumps between successive segments. Note that in the case of real world data, we actually adopt an existing state-of-the-art open-source SLAM framework~\cite{campos2021orb}. However, the standard framework does not concatenate successive trajectory segments but---upon tracking and re-localization failure---always initializes new maps from the same global reference frame (often set to identity rotation and zero translation). Hence, it would fail to properly initialize our hybrid pose graph optimizer.

Throughout our experiments, we compare the following results:
\begin{itemize}
  \item NO-PGO: Simple concatenation of successive trajectory segments, but no loop closure and pose graph optimization.
  \item SPGO: Scale-drift aware pose graph optimization.
  \item HPGO: Scale-drift and scale-jump aware pose graph optimization.
  \item GT: Ground truth obtained from external motion tracking system or wheel odometry (for qualitative comparison only).
\end{itemize}

Given that in the monocular case an overall scale ambiguity will always remain, the final trajectories are aligned by SIM3 alignment using Umeyama's method \cite{umeyama1991least} applied to the first 20 keyframes of each trajectory. We use the \textit{evo} odometry evaluation tool \cite{grupp2017evo} to perform the alignment and compare the results obtained by the different methods.

\subsection{Analysis on simulated data}

In order to validate the cases in which global scale consistency can be re-installed, we start by examining simulated planar trajectories during which the camera alternates between pure forward motion along the principal axis and pure rotations about the vertical $y$-axis, the idea being that the robot loses track of the local map when it rotates and the scale jumps after re-initialization. In the first experiment, the camera exhibits four equally long segments and rotates 120 degrees between successive segments. Each scale-consistent trajectory segment is composed of 20 key frames, and the assigned scales to each segment are 1, 0.5, 2, and 0.3. As a result of the rotations, the real trajectory admits a triangular shape and the final segment overlaps with the first one, thus simulating a scenario in which a loop closure occurs. The assumption therefore is that the relative scale between segments one and two, two and three, and three and four are unknown, while the relative scale between the last and the first segment is measured. As indicated in Figure~\ref{fig:simulation_traj}~(a), the simple concatenation of the scaled segments leads to an obviously wrong result. SPGO on the other hand only models potential drift in the scale variable, but not discontinuous behavior. SPGO thus attempts to distribute the scale error uniformly over the whole graph, which is why the optimization result still presents severe distortions. As expected, our proposed HPGO framework recovers the exact ground truth trajectory.

In order to verify the importance of the number of critical nodes along the loop, in a next experiment we augment the number of segments to 5 and reduce the inter-segment rotation angles to 90 degrees. The real trajectory therefore becomes a square, and it is now the fifth segment that overlaps with the first one. The assigned scales are 1, 0.5, 2, 0.2, and 0.8. As can be observed in Figure~\ref{fig:simulation_traj}~(b), the simple concatenation of the scaled estimates again leads to severe drift, and SPGO is again unable to recover a correct, distortion-free result. Though better, our proposed HPGO this time also fails to recover a distortion-free result, and returns a rectangular trajectory.

To confirm the importance of the spatial distribution of critical nodes rather than the exact shape of the trajectory, a third experiment was conducted. The camera moved along a flat circular path, with the principal axis tangential to the trajectory. In a first try we let the camera move along four thirds of a circle, again assuming critical nodes between segment one and two, two and three, and three and four. The fourth segment is again overlapping with the first, thus simulating the loop closure. As indicated in Figure~\ref{fig:simulation_traj}~(c), the result is again similar to the case of the triangular trajectory, and HPGO is the only method able to recover an undistorted, globally consistent trajectory. On the other hand, as depicted in Figure~\ref{fig:simulation_traj}~(d), if the camera exhibits five quarters of a circle, and the number of critical nodes again augments to four, the final result remains distorted. Note that numerical results for all simulations are indicated in Table~\ref{table:ateSimulation}.

\begin{table}[]
    \caption{Average Trajectory Error (ATE) for all Simulation Results}
\resizebox{\linewidth}{!}{
    \begin{tabular}{|c|c|ccccc|}
    \hline
    \multirow{2}{*}{Dataset}   & \multirow{2}{*}{Method} & \multicolumn{5}{c|}{ATE(m)}                                                                                                         \\ \cline{3-7} 
                               &                         & \multicolumn{1}{c|}{rmse}   & \multicolumn{1}{c|}{mean}   & \multicolumn{1}{c|}{median} & \multicolumn{1}{c|}{std}    & sse      \\ \hline\hline
    \multirow{3}{*}{Triangle}  & HPGO                    & \multicolumn{1}{c|}{\textbf{$<$10e-6}} & \multicolumn{1}{c|}{\textbf{$<$10e-6}} & \multicolumn{1}{c|}{\textbf{$<$10e-6}} & \multicolumn{1}{c|}{\textbf{$<$10e-6}} & \textbf{$<$10e-6}   \\ \cline{2-7} 
                               & NO-PGO                  & \multicolumn{1}{c|}{3.475}  & \multicolumn{1}{c|}{2.575}  & \multicolumn{1}{c|}{2.000}  & \multicolumn{1}{c|}{2.334}  & 1014.552 \\ \cline{2-7} 
                               & SPGO                    & \multicolumn{1}{c|}{1.922}  & \multicolumn{1}{c|}{1.657}  & \multicolumn{1}{c|}{1.354}  & \multicolumn{1}{c|}{0.973}  & 310.245  \\ \hline\hline
    \multirow{3}{*}{Rectangle} & HPGO                    & \multicolumn{1}{c|}{\textbf{1.528}}  & \multicolumn{1}{c|}{\textbf{1.054}}  & \multicolumn{1}{c|}{\textbf{0.659}}  & \multicolumn{1}{c|}{\textbf{1.107}}  & \textbf{245.352}  \\ \cline{2-7} 
                               & NO-PGO                  & \multicolumn{1}{c|}{3.124}  & \multicolumn{1}{c|}{2.533}  & \multicolumn{1}{c|}{2.828}  & \multicolumn{1}{c|}{1.828}  & 1024.604 \\ \cline{2-7} 
                               & SPGO                    & \multicolumn{1}{c|}{1.741}  & \multicolumn{1}{c|}{1.316}  & \multicolumn{1}{c|}{0.902}  & \multicolumn{1}{c|}{1.139}  & 318.172  \\ \hline\hline
    \multirow{3}{*}{Circle4}   & HPGO                    & \multicolumn{1}{c|}{\textbf{$<$10e-6}} & \multicolumn{1}{c|}{\textbf{$<$10e-6}} & \multicolumn{1}{c|}{\textbf{$<$10e-6}} & \multicolumn{1}{c|}{\textbf{$<$10e-6}} & \textbf{$<$10e-6}   \\ \cline{2-7} 
                               & NO-PGO                  & \multicolumn{1}{c|}{4.589}  & \multicolumn{1}{c|}{3.485}  & \multicolumn{1}{c|}{3.054}  & \multicolumn{1}{c|}{2.985}  & 1705.623 \\ \cline{2-7} 
                               & SPGO                    & \multicolumn{1}{c|}{2.276}  & \multicolumn{1}{c|}{1.710}  & \multicolumn{1}{c|}{1.357}  & \multicolumn{1}{c|}{1.502}  & 419.459  \\ \hline\hline
    \multirow{3}{*}{Circle5}   & HPGO                    & \multicolumn{1}{c|}{\textbf{2.159}}  & \multicolumn{1}{c|}{\textbf{1.502}}  & \multicolumn{1}{c|}{\textbf{1.019}}  & \multicolumn{1}{c|}{\textbf{1.551}}  & \textbf{470.844}  \\ \cline{2-7} 
                               & NO-PGO                  & \multicolumn{1}{c|}{4.883}  & \multicolumn{1}{c|}{3.953}  & \multicolumn{1}{c|}{4.097}  & \multicolumn{1}{c|}{2.868}  & 2408.688 \\ \cline{2-7} 
                               & SPGO                    & \multicolumn{1}{c|}{2.352}  & \multicolumn{1}{c|}{1.734}  & \multicolumn{1}{c|}{1.148}  & \multicolumn{1}{c|}{1.589}  & 558.486  \\ \hline
    \end{tabular}}
    \label{table:ateSimulation}
    \end{table}

\subsection{Validation on real data}

In our next experiment, we apply our proposed method to an existing Monocular SLAM system: ORB-SLAM3~\cite{campos2021orb}. The latter is a complete, modern system capable of handling tracking failures and multiple maps using Atlas\cite{elvira2019orbslam}.
To make our method work, we must adapt the bootstrapping and loop closure modules. The new bootstrapping module is designed for small dual-drive platforms capable of pure rotational displacements. First, we disable relocalization attempts after a tracking failure, and immediately attempt re-initizalition. We also detect pure rotations, and only bootstrap the estimation if sufficient displacement occurs. We furthermore use 1-point Ransac \cite{scaramuzza20111} to robustly initialize the algorithm even if only small displacements occur. After re-initialization, the edge connecting the last frame of the previous segment and the first frame of the new segment has pure rotation and unknown scale. Finally, the loop closure module is modified to use HPGO, and fragmented maps are merged into a complete and scale-consistent map after optimization.
\begin{figure}[t]
    \centering
    \subfigure[3-Bar]{
        \begin{minipage}[t]{0.49\linewidth}
            \centering
            \includegraphics[width=\linewidth]{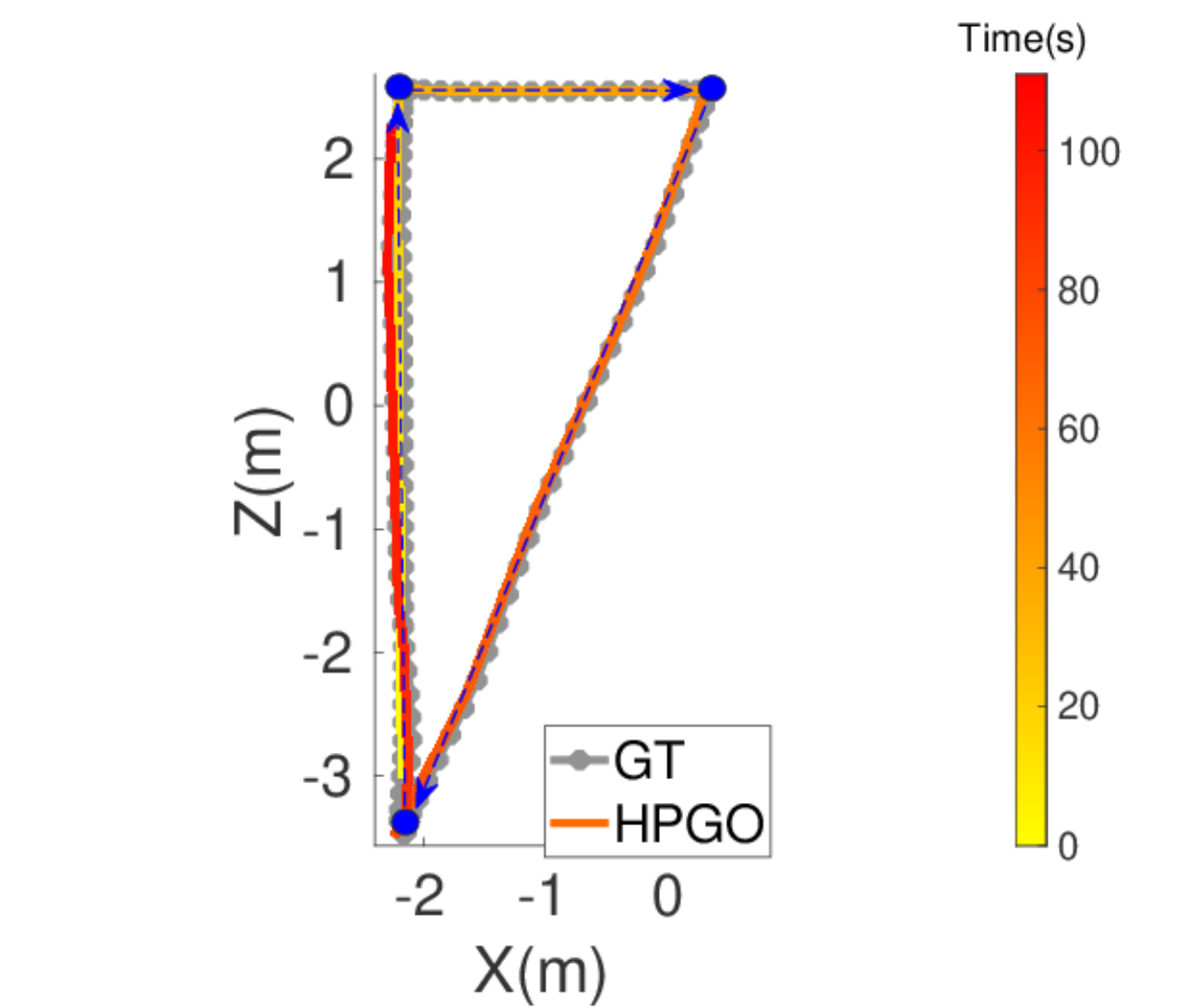}
            \\
            \includegraphics[width=\linewidth]{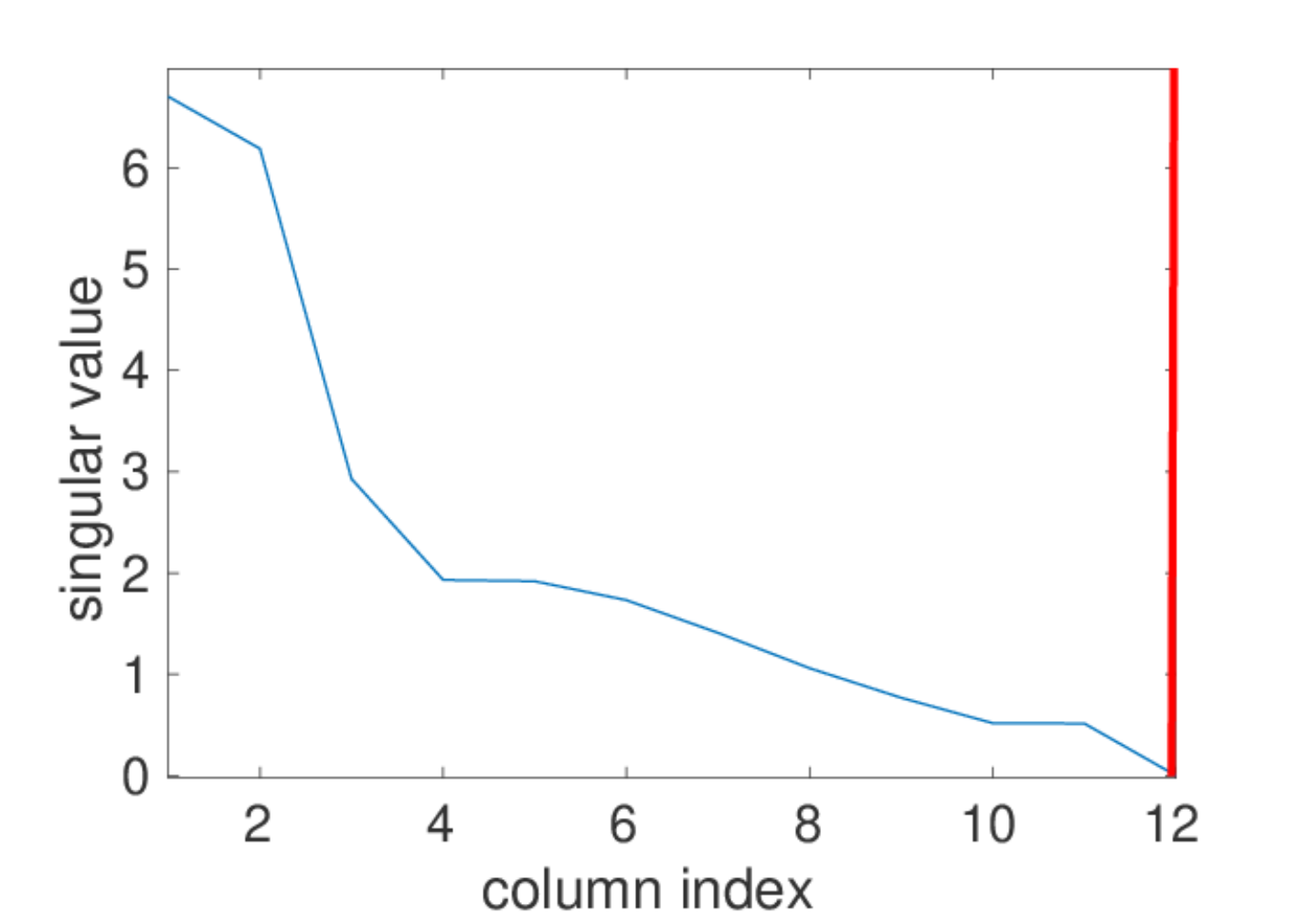}
        \end{minipage}%

    }%
    \subfigure[4-Bar]{
        \begin{minipage}[t]{0.49\linewidth}
            \centering
            \includegraphics[width=\linewidth]{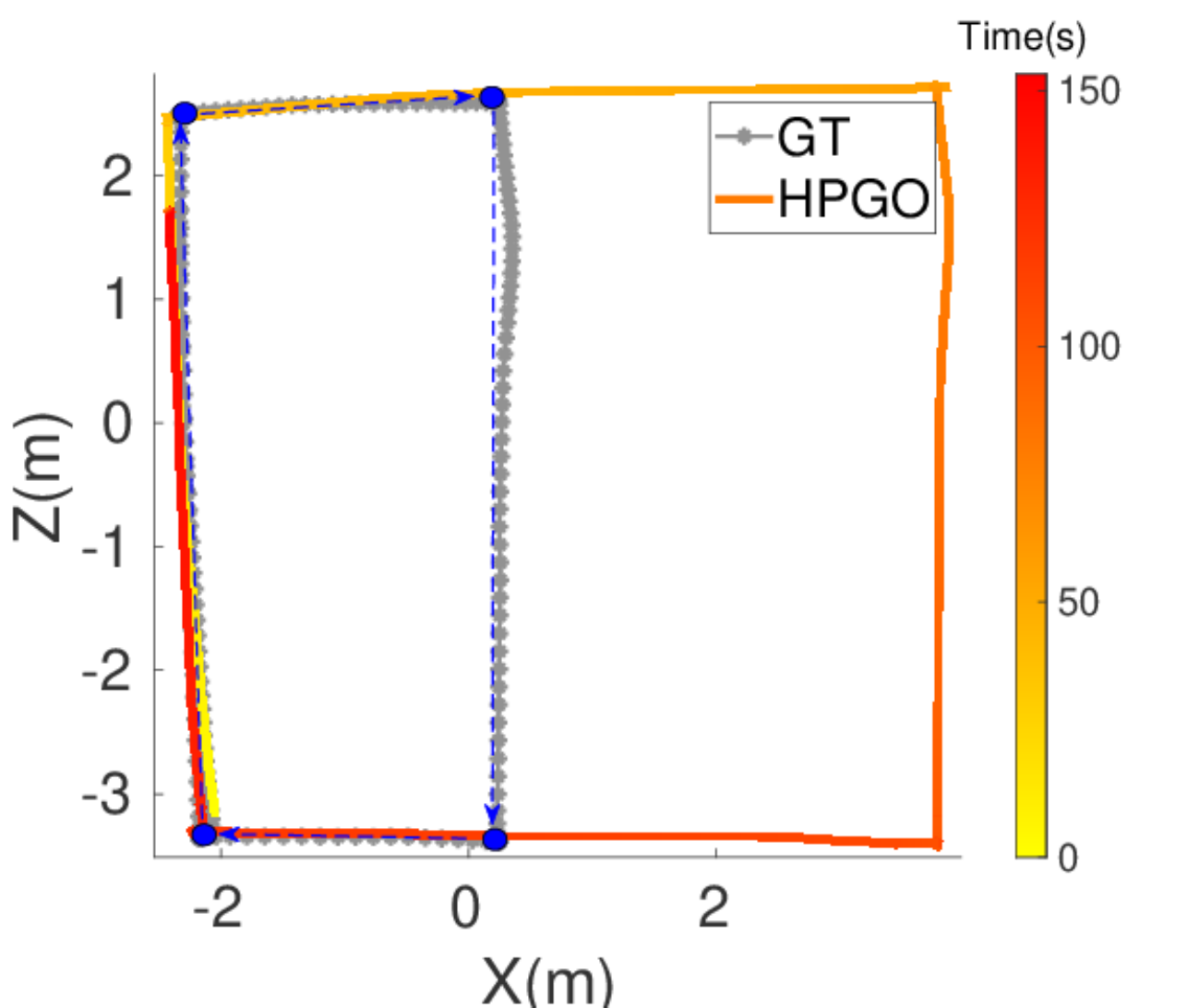}
            \\
            \includegraphics[width=\linewidth]{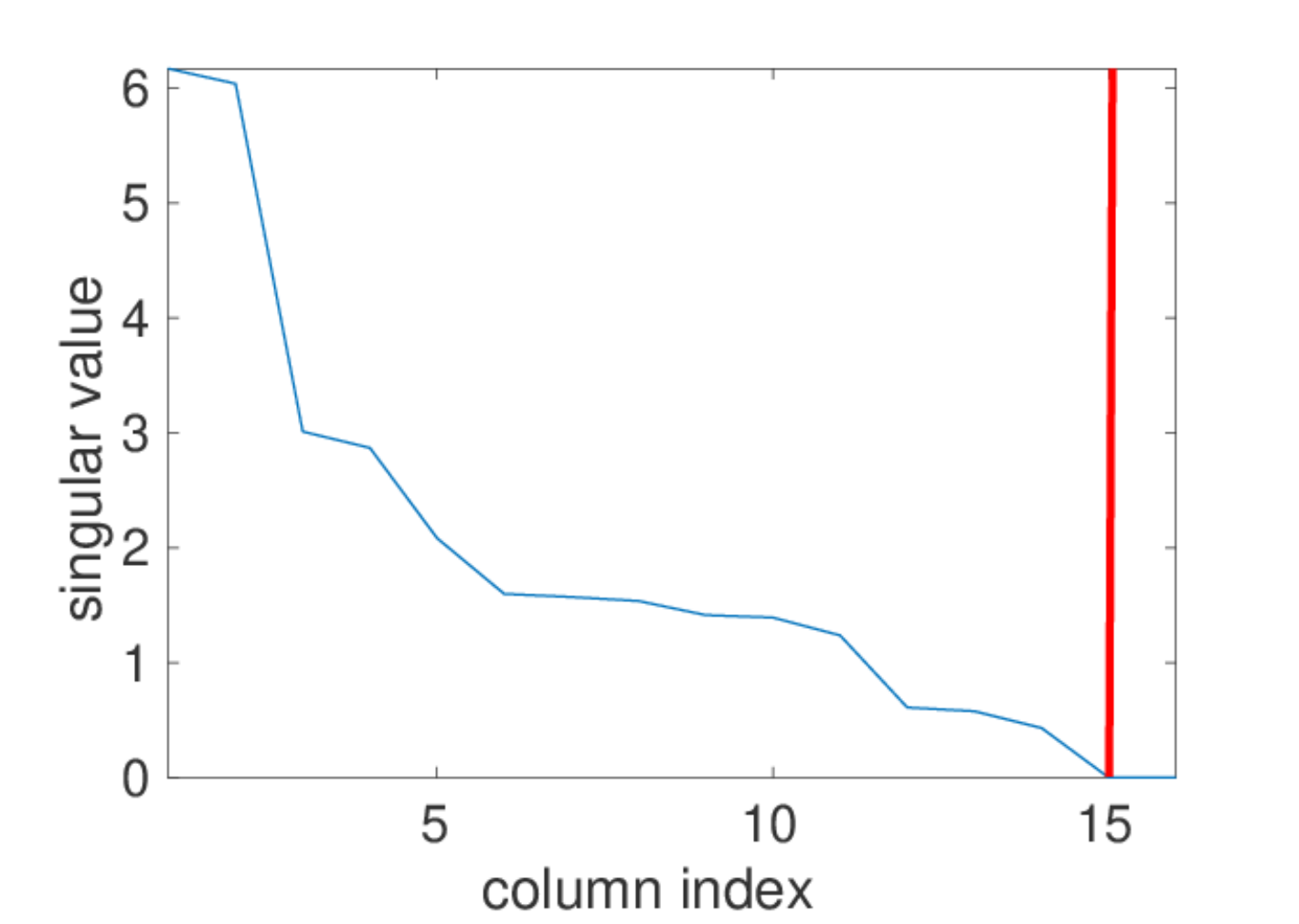}
        \end{minipage}
    }%

\centering
\caption{Trajectory and rank evaluation of real-world cases with three or four critical nodes. Top row: bird-eye view onto the estimated trajectories including a comparison against groundtruth (obtained by mocap system). Blue points denote critical nodes, while directional vectors indicate the connections between critical nodes. Bottom row: ordered singular values of matrix $\mathbf{A}$ (from largest to smallest). The red line marks the break-down point where the singular values become very small.}
\label{fig:traj_and_rank}
\end{figure}

\begin{table}[h]
\caption{Absolute Trajectory Error (ATE) on Real-world Sequences}
\resizebox{\linewidth}{!}{
\begin{tabular}{|c|c|ccccc|}
\hline
\multirow{2}{*}{Dataset} & \multirow{2}{*}{Method} & \multicolumn{5}{c|}{ATE(m)}                                                                                                                                                 \\ \cline{3-7} 
                         &                         & \multicolumn{1}{c|}{rmse}           & \multicolumn{1}{c|}{mean}           & \multicolumn{1}{c|}{median}         & \multicolumn{1}{c|}{std}            & sse              \\ \hline\hline
\multirow{3}{*}{seq01}   & HPGO                    & \multicolumn{1}{c|}{\textbf{0.520}} & \multicolumn{1}{c|}{\textbf{0.444}} & \multicolumn{1}{c|}{\textbf{0.437}} & \multicolumn{1}{c|}{\textbf{0.270}} & \textbf{66.294}  \\ \cline{2-7} 
                         & NO-PGO                  & \multicolumn{1}{c|}{0.904}          & \multicolumn{1}{c|}{0.759}          & \multicolumn{1}{c|}{0.676}          & \multicolumn{1}{c|}{0.491}          & 238.158          \\ \cline{2-7} 
                         & SPGO                    & \multicolumn{1}{c|}{1.973}          & \multicolumn{1}{c|}{1.544}          & \multicolumn{1}{c|}{1.619}          & \multicolumn{1}{c|}{1.229}          & 1012.624         \\ \hline\hline
\multirow{3}{*}{seq02}   & HPGO                    & \multicolumn{1}{c|}{\textbf{0.248}} & \multicolumn{1}{c|}{\textbf{0.198}} & \multicolumn{1}{c|}{\textbf{0.148}} & \multicolumn{1}{c|}{\textbf{0.150}} & \textbf{24.359}  \\ \cline{2-7} 
                         & NO-PGO                  & \multicolumn{1}{c|}{5.339}          & \multicolumn{1}{c|}{4.619}          & \multicolumn{1}{c|}{4.975}          & \multicolumn{1}{c|}{2.678}          & 13715.348        \\ \cline{2-7} 
                         & SPGO                    & \multicolumn{1}{c|}{1.613}          & \multicolumn{1}{c|}{1.402}          & \multicolumn{1}{c|}{1.502}          & \multicolumn{1}{c|}{0.796}          & 1103.409         \\ \hline\hline
\multirow{3}{*}{seq03}   & HPGO                    & \multicolumn{1}{c|}{\textbf{0.054}} & \multicolumn{1}{c|}{\textbf{0.048}} & \multicolumn{1}{c|}{\textbf{0.046}} & \multicolumn{1}{c|}{\textbf{0.023}} & \textbf{0.484}   \\ \cline{2-7} 
                         & NO-PGO                  & \multicolumn{1}{c|}{2.943}          & \multicolumn{1}{c|}{2.671}          & \multicolumn{1}{c|}{3.172}          & \multicolumn{1}{c|}{1.236}          & 2149.201         \\ \cline{2-7} 
                         & SPGO                    & \multicolumn{1}{c|}{0.889}          & \multicolumn{1}{c|}{0.803}          & \multicolumn{1}{c|}{0.827}          & \multicolumn{1}{c|}{0.382}          & 133.810          \\ \hline\hline
\multirow{3}{*}{seq04}   & HPGO                    & \multicolumn{1}{c|}{\textbf{0.092}} & \multicolumn{1}{c|}{\textbf{0.082}} & \multicolumn{1}{c|}{\textbf{0.074}} & \multicolumn{1}{c|}{\textbf{0.040}} & \textbf{1.335}   \\ \cline{2-7} 
                         & NO-PGO                  & \multicolumn{1}{c|}{1.078}          & \multicolumn{1}{c|}{0.907}          & \multicolumn{1}{c|}{1.130}          & \multicolumn{1}{c|}{0.583}          & 211.767          \\ \cline{2-7} 
                         & SPGO                    & \multicolumn{1}{c|}{0.355}          & \multicolumn{1}{c|}{0.292}          & \multicolumn{1}{c|}{0.238}          & \multicolumn{1}{c|}{0.201}          & 19.310           \\ \hline\hline
\multirow{3}{*}{seq05}   & HPGO                    & \multicolumn{1}{c|}{\textbf{0.032}} & \multicolumn{1}{c|}{\textbf{0.029}} & \multicolumn{1}{c|}{\textbf{0.026}} & \multicolumn{1}{c|}{\textbf{0.013}} & \textbf{0.058}   \\ \cline{2-7} 
                         & NO-PGO                  & \multicolumn{1}{c|}{1.378}          & \multicolumn{1}{c|}{1.342}          & \multicolumn{1}{c|}{1.286}          & \multicolumn{1}{c|}{0.313}          & 169.115          \\ \cline{2-7} 
                         & SPGO                    & \multicolumn{1}{c|}{0.654}          & \multicolumn{1}{c|}{0.607}          & \multicolumn{1}{c|}{0.532}          & \multicolumn{1}{c|}{0.243}          & 36.414           \\ \hline\hline
\multirow{3}{*}{seq06}   & HPGO                    & \multicolumn{1}{c|}{\textbf{1.081}} & \multicolumn{1}{c|}{\textbf{1.065}} & \multicolumn{1}{c|}{\textbf{1.075}} & \multicolumn{1}{c|}{\textbf{0.182}} & \textbf{107.619} \\ \cline{2-7} 
                         & NO-PGO                  & \multicolumn{1}{c|}{1.828}          & \multicolumn{1}{c|}{1.738}          & \multicolumn{1}{c|}{1.624}          & \multicolumn{1}{c|}{0.566}          & 337.604          \\ \cline{2-7} 
                         & SPGO                    & \multicolumn{1}{c|}{1.455}          & \multicolumn{1}{c|}{1.364}          & \multicolumn{1}{c|}{1.542}          & \multicolumn{1}{c|}{0.505}          & 182.197          \\ \hline
\end{tabular}
}
\label{table:ateWithGT}
\end{table}

We apply our HPGO-based ORB-SLAM3 system to real-world data captured by the kobuki robot. The robot is capable of executing forward motion and pure rotations. Images are captured at 30 FPS using an onboard RGB camera with a resolution of $640 \times 480$ pix and a horizontal FoV of 57$^{\circ}$ and a vertical FoV of 43 $^{\circ}$. We use the OptiTrack motion capture system to capture the ground truth trajectory of the robot. As shown in Fig. \ref{fig:traj_and_rank} (a), our HPGO-based system can reconcile a globally consistent result in a situation with three re-initializations, but fails to reconcile scales in a scenario with 4 tracking failures (cf. Fig.\ref{fig:traj_and_rank} (b)). The bottom row additionally illustrates the singular values of matrix $\mathbf{A}$ introduced in (\ref{equ:evaluation_matrix}). As can be observed, the matrix has additional rank defficiencies in the latter under-constrained case, which confirms its potential use in analysing the quality of the global scale reconciliation at the end of the optimization. Full ATE results are again listed in Table.~\ref{table:ateWithGT}.

\subsection{Tests on real-world, room-size scenarios}

For the final experiment, in order to evaluate the practicality of our HPGO-based ORB-SLAM3 system, the robot is assigned an exploratory task in a regular indoor environment (e.g. an office room). We use the onboard wheel odometry signals to obtain a metrically scaled, globally consistent version of the trajectory. It may not be as accurate as mocap ground truth, but still serves well towards a qualitative analysis of the algorithm's scale reconciliation ability.
\begin{figure}[h]
    \centering
    \subfigure[single loop]{
        \begin{minipage}[t]{0.49\linewidth}
            \centering
            \includegraphics[width=\linewidth]{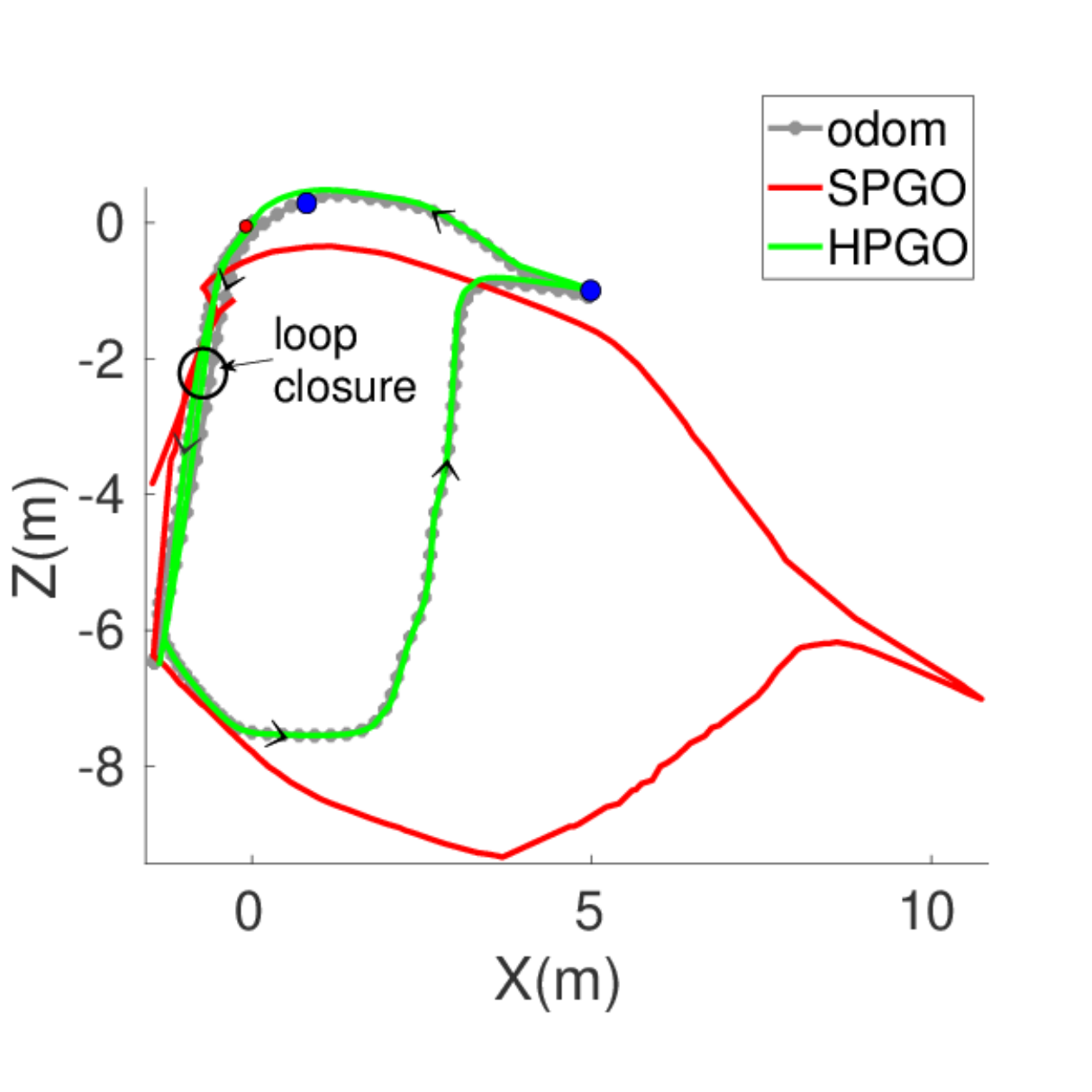}
            \\
            \includegraphics[width=\linewidth]{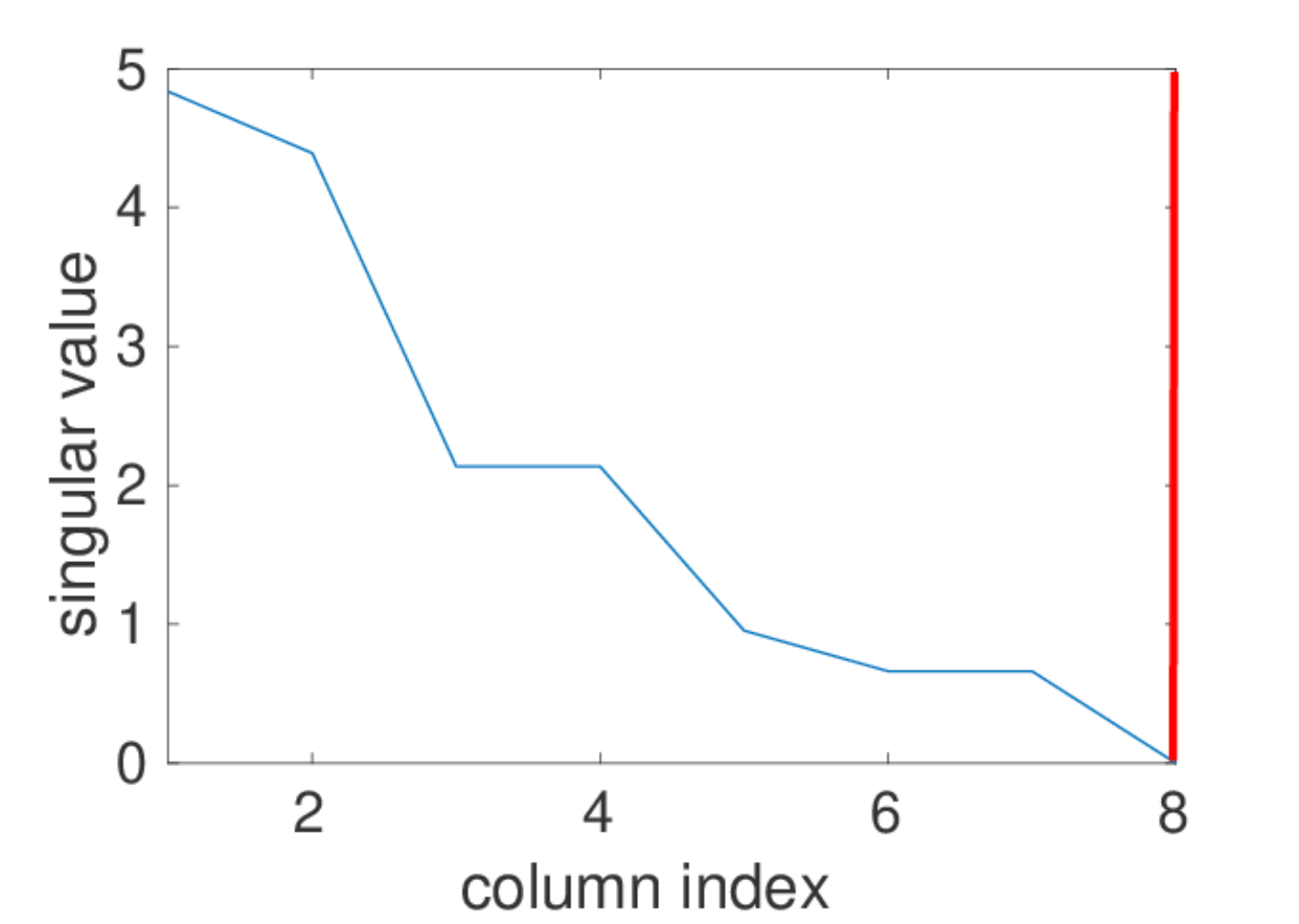}
        \end{minipage}%

    }%
    \subfigure[two loops]{
        \begin{minipage}[t]{0.49\linewidth}
            \centering
            \includegraphics[width=\linewidth]{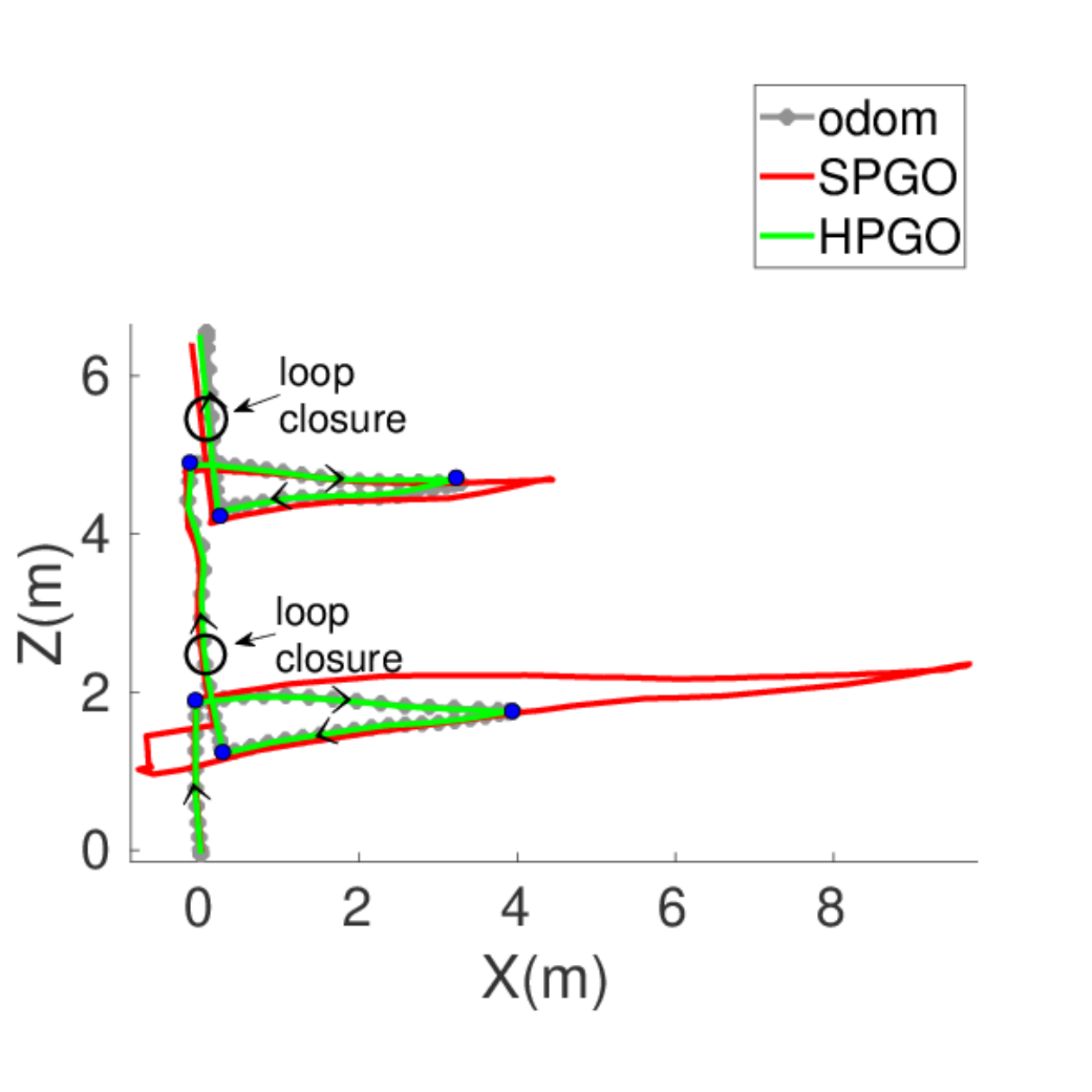}
            \\
            \includegraphics[width=\linewidth]{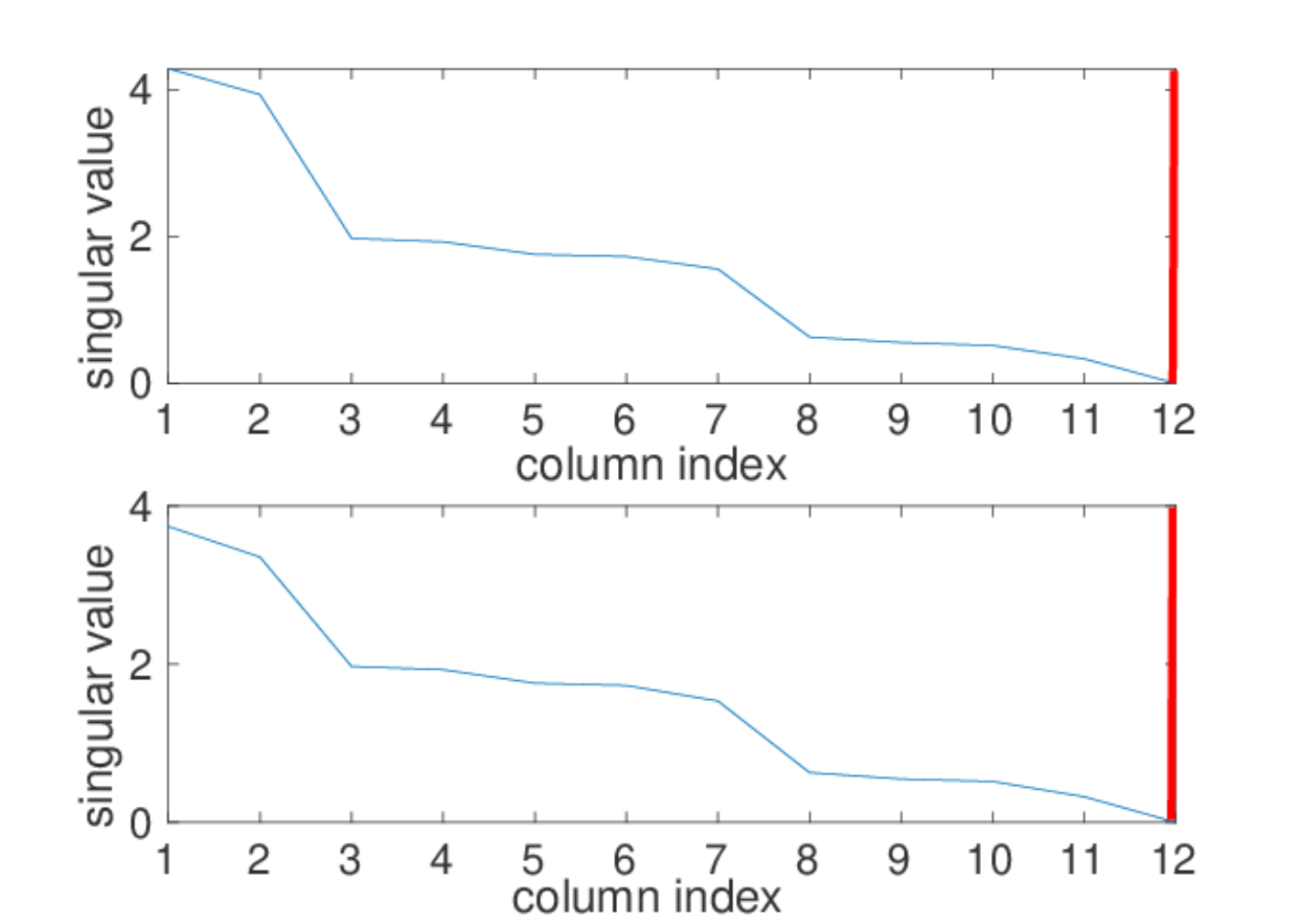}
        \end{minipage}
    }%

\centering
\caption{Trajectory and rank evaluation of room-size real-world cases. Top row: bird-eye view onto the estimated trajectories with comparison against odometry and SPGO. Blue points denote critical nodes, red points indicate the initial position, and arrows show movement direction. Bottom row: ordered singular values of matrix $\mathbf{A}$ (from largest to smallest) with break-down mark for each loop closure. For (b), the top and bottom figures correspond to the situation after the first and second loop closure, respectively.}
\label{fig:no_gt_traj}
\end{figure}

As shown in Figure~\ref{fig:no_gt_traj} (a), in a 2-Bar constrained scenario, our HPGO-based system reconciles a globally consistent result while SPGO leads to severe distortions. Figure~\ref{fig:no_gt_traj} (b) illustrates the case of executing HPGO multiple times and incrementally constructing a scale-consistent trajectory despite 6 tracking failures and re-initializations occurring along the loop. A video demonstrating this result in real-time is indicated in the supplementary material.

\section{CONCLUSION}
\label{sec:conclusion}

Our proposed hybrid pose graph optimization method represents a transparent way to reconcile multiple individually scaled trajectory segments captured by a monocular camera through a single optimization framework. The method is able to reconstruct accurate global trajectory results despite multiple re-initializations occurring along the loop. Our contribution furthermore comprises a theoretical analysis of global scale consistency purely based on the number, connectivity, and spatial arrangement of the re-initialization nodes. Although most results are obtained for a planar ground vehicle platform, it is worth noting that the optimization in all experiments was indeed carried out over SIM3. The complete theory presented in the paper remains valid for the general case. We believe that our contribution is of interest to the community, and plan to release a patch for ORB-SLAM3 to enable re-usability.

\section{Acknowledgement}
The research presented in this paper has been supported by projects 22dz1201900 and 22ZR1441300 funded by the Shanghai Science Foundation as well as the project 62250610225 by the National Science Foundation of China. The authors would furthermore like to acknowledge the support provided by Midea Robozone.
\bibliographystyle{IEEEtran}
\bibliography{IEEEexample}
\end{document}